\documentclass[10pt,journal,compsoc]{IEEEtran}

\usepackage{cite}
\usepackage{amsmath}
\usepackage{hyperref}
\hypersetup{
	colorlinks=true, linkcolor=blue, citecolor=green, urlcolor=magenta
}
\usepackage{amssymb}
\usepackage{booktabs}
\usepackage{array}
\usepackage[caption=false,font=footnotesize]{subfig}
\usepackage{fixltx2e}
\usepackage{stfloats}
\usepackage{url}
\usepackage{graphicx}
\usepackage{multirow}
\usepackage{todonotes}
\usepackage{diagbox}
\usepackage[table]{xcolor}

\begin{document}
%

\title{MITE-Net: SWaP-Optimized 4K Video Tiny Target Perception for Embodied Edge SAR}

\author{
	 Mingshuo~Xu\textsuperscript{\textdagger},~\IEEEmembership{Student~Member,~IEEE},
	 Mu~Hua\textsuperscript{\textdagger},
     Jigen~Peng,
     Qi~Wang\textsuperscript{\textasteriskcentered},
     and~Shigang~Yue\textsuperscript{\textasteriskcentered},~\IEEEmembership{Senior~Member,~IEEE}
	\thanks{Mingshuo Xu, Mu Hua, Qi Wang, and Shigang Yue were with the School of Mathematics and Computing Science, University of Leicester, Leicester LE1 7RH, UK. E-mail: \{mx60, mh744, qw96, sy237\}@leicester.ac.uk. }
	\thanks{Jigen Peng was with the Machine Life and Intelligence Research Center, Guangzhou University, Guangzhou 510006, China. E-mail: jgpeng@gzhu.edu.cn.}

	\thanks{The SAR-Tiny datasets and source code are continuously being updated in \href{https://github.com/MingshuoXu/MITE-Net.git}{Project Repository} and  \href{https://github.com/MingshuoXu/SAR-Tiny.git}{Dataset Repository}.}

	\thanks{\textsuperscript{\textdagger}Mingshuo Xu and Mu Hua contributed equally to this work.}
	\thanks{\textsuperscript{\textasteriskcentered}Corresponding author: Qi Wang and Shigang Yue.}
	}

\IEEEtitleabstractindextext{%
\begin{abstract}
Real-time tiny target perception in high-resolution imagery is critical for embodied Search-and-Rescue (SAR) missions. However, strict Size, Weight, and Power (SWaP) constraints on edge devices like UAVs create a bottleneck: traditional image downsampling causes severe feature loss, while slice-based processing incurs prohibitive latency. To address this gap, this paper introduces a comprehensive framework encompassing a novel architecture, specialized datasets, and hardware-level benchmarks. First, we propose MITE-Net, a SWaP-optimized cascaded architecture, which couples a bio-inspired, learning-free Tiny Target Motion-Based Region Proposal Network (TTM-RPN) with a sub-0.14M-parameter R-CNN-like head. Second, to standardize 4K tiny target evaluation, we construct the SAR-Tiny Datasets by relabeling two challenging UAV datasets: SeaDroneSee-Tiny (dynamic maritime scenes, tiny targets predominantly of 64–256 pixels ) and UAVID-Tiny (cluttered urban scenes, extremely tiny targets, $\le$ 64 pixels). Third, we benchmark against state-of-the-art YOLO models on an edge device, NVIDIA Jetson AGX Xavier, where MITE-Net directly processes 4K maritime imagery, achieving a 100\% search success rate at 30.33 FPS. Consuming merely 3.19 W (9.51 FPS/W), MITE-Net vastly outperforms YOLO baselines in target recall and energy efficiency. Conversely, UAVID-Tiny evaluations expose a compound structural limitation: the learning-free bionic front-end struggles against urban backgrounds, while the ultra-lightweight head lacks representational capacity for complex features. Ultimately, this work delivers an efficient onboard perception paradigm and a rigorous baseline guiding future end-to-end SAR architectures. 
\end{abstract}

}

\maketitle

\IEEEdisplaynontitleabstractindextext

%


\IEEEraisesectionheading{\section{Introduction}\label{Sect_Intro}}

\IEEEPARstart{S}{earch-and-rescue} (SAR) and emergency response missions increasingly require Unmanned Aerial Vehicles (UAVs) to perceive, localize, and respond to critical events in real time \cite{erdelj2017help}. In such scenarios, the early discovery of tiny targets from high-resolution imagery is often mission-critical. Typical examples include stranded individuals in maritime SAR, small ignition sources in wildfire monitoring, and hazardous objects in disaster assessment \cite{wang2021tiny, cheng2023towards}. 

\begin{figure}[!t]
	\centering
	\includegraphics[width=0.48\textwidth]{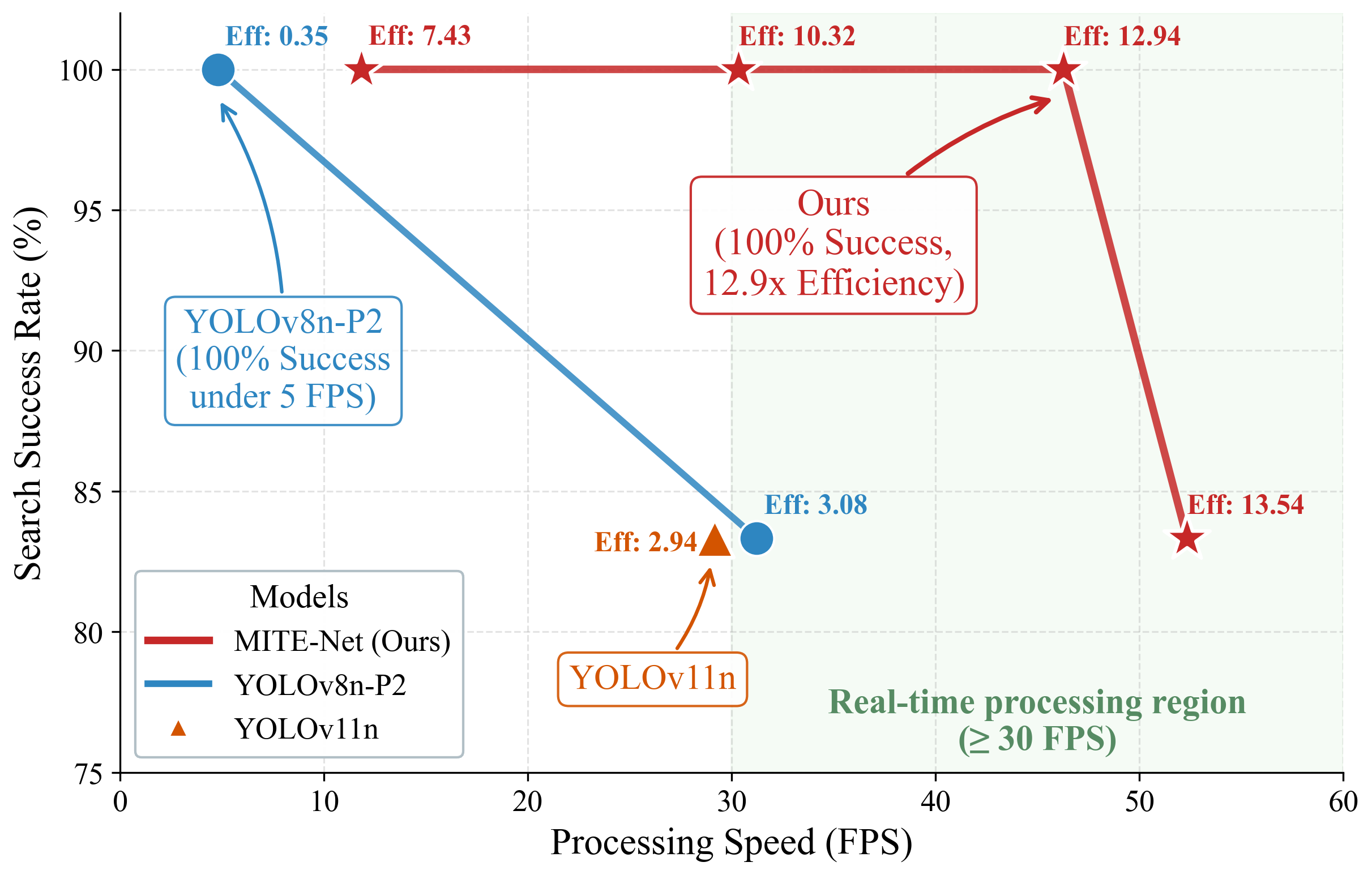}
	\caption{Comparison of search success rate versus processing speed (FPS) among different models. The green shaded area denotes the real-time processing region ($\geq$ 30 FPS), and the annotated numbers indicate the efficiency metric for each tested configuration. While YOLOv8n-P2 achieves a 100\% success rate, it is limited to suboptimal speeds ($<$ 5 FPS). Conversely, YOLOv11n approaches real-time speed but suffers a significant drop in success rate. Our proposed MITE-Net achieves an optimal trade-off, successfully maintaining a 100\% search success rate within the real-time processing region while demonstrating significantly higher efficiency (12.9$\times$).}
	\label{Fig_Efficiency}
\end{figure}

However, high-resolution tiny-target perception is computationally demanding in practical applications such as embedded systems, mobile platforms, and remote sensing, particularly for wide-area 4K aerial imagery. Under strict Size, Weight, and Power (SWaP) constraints, the challenge is further compounded by the scarce spatial scale of the tiny targets, which may span only a few tens of pixels even in 4K imagery. Consequently, conventional detectors face a fundamental perception bottleneck: direct downsampling improves throughput but destroys fine-grained target cues, whereas slice-based inference preserves local detail but incurs prohibitive computational overhead and latency (see Fig.~\ref{Fig_Efficiency}). Ultimately, existing lightweight detectors struggle to balance target recall, real-time execution, and energy efficiency in edge SAR settings.

To address this gap and enable true embodied edge intelligence, this paper introduces a comprehensive framework encompassing a novel architecture, specialized datasets, and hardware-level benchmarks. First, we propose \textbf{MITE-Net} (Motion-Informed Tiny-target Edge Network), a SWaP-optimized cascaded architecture specifically designed for efficient onboard perception. To overcome the fundamental feature loss dilemma associated with standard downsampling, MITE-Net uniquely decouples the spatial perception pipeline. Specifically, it utilizes a learning-free Tiny Target Motion-based Region Proposal Network (TTM-RPN), built upon the bio-inspired various-velocity Small Target Motion Detector (vSTMD) \cite{xu2025vstmd}, which operates in a downsampled space to rapidly mine spatiotemporal motion cues at a very low computational cost. Subsequently, to avoid any loss of microscopic details, a sparse set of motion-informed Regions of Interest (RoIs) is cropped directly from the raw 4K imagery. These native-resolution RoIs are then forwarded to a highly tailored, sub-0.14M-parameter R-CNN-like head for precise classification and localization. This mechanism successfully bridges the gap between high-frequency feature preservation and real-time inference, enabling highly sensitive target perception under strict edge SWaP constraints.

\begin{figure*}[!t]
	\centering
	\includegraphics[width=1\textwidth]{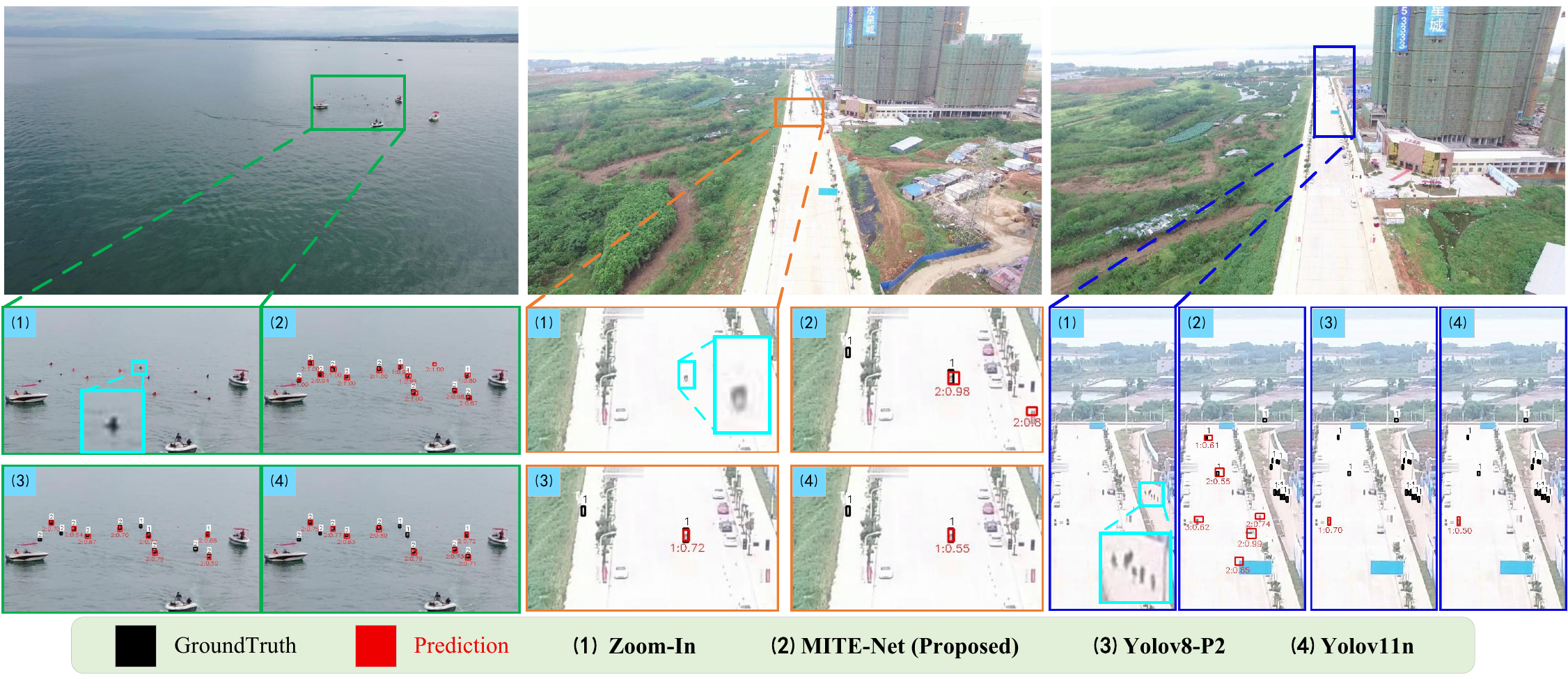}
	\caption{Qualitative demonstration of 4K tiny target perception for the SAE mission using the proposed MITE-Net and baseline models. The top row displays the original 4K resolution images, while the bottom panels show the corresponding zoomed-in regions. As observed, due to the extremely tiny scale of the targets, they appear merely as blurry spots even after significant magnification. The first column presents results on the SeaDroneSee-Tiny dataset, where our proposed model demonstrates excellent detection performance. The second and third columns exhibit results on the UAVID-Tiny dataset. Specifically, in the second column, our model successfully localizes a target but suffers from a classification failure. In the third column, our model experiences a bounding box regression failure when detecting multiple tiny objects. Notably, when these targets consist of merely a dozen pixels, the recall rate of mainstream YOLO baselines also drops to near zero.}
	\label{Fig_Output}
\end{figure*}

Second, we construct and open-source the 4K SAR-Tiny datasets to rigorously standardize 4K tiny target evaluation under realistic edge SAR conditions. By relabelling two prominent 4K UAV datasets \cite{kiefer20221st, varga2022seadronessee, lyu2020uavid}, we define two tiers of escalating difficulty: SeaDroneSee-Tiny for dynamic maritime environments (tiny targets predominantly 64–256 pixels), and UAVID-Tiny for highly cluttered urban environments (extremely tiny targets, $\le$ 64 pixels). Leveraging these datasets, we conduct extensive edge-hardware benchmarking against state-of-the-art baselines, including YOLOv11n and YOLOv8n-P2, to rigorously evaluate MITE-Net's performance and validate its superiority under realistic SWaP bottlenecks, with representative visual comparisons shown in Fig.~\ref{Fig_Output}.

In summary, the main contributions of this work are as follows:
\begin{itemize}
	\item \textbf{A SWaP-Optimized Edge Perception Architecture}: The proposed MITE-Net decouples spatial perception to overcome the feature loss dilemma. By running a bionic motion front-end in a downsampled space and cropping RoIs from raw 4K imagery for a highly tailored, $<$0.14M parameter head, it enables real-time, high-resolution perception under strict edge constraints.
	
	\item \textbf{SAR-Tiny Datasets and 4K Edge Benchmarks}: We construct and open-source the SAR-Tiny Datasets (derived from SeaDroneSee \cite{kiefer20221st, varga2022seadronessee} and UAVID\cite{lyu2020uavid}) to standardize 4K tiny target evaluation. Alongside this, we provide rigorous edge-hardware benchmarks against state-of-the-art YOLO models (YOLOv11n, YOLOv8n-P2), establishing critical baselines.
	 
	\item \textbf{Hardware-Level Efficiency and Structural Boundaries}: Deployed on an edge device, NVIDIA Jetson AGX Xavier, MITE-Net achieves 30.33 FPS and a 100\% target recall at merely 3.19 W (9.51 FPS/Watt), vastly outperforming YOLO baselines. Furthermore, we candidly map the representational boundaries of ultra-lightweight models in hyper-cluttered urban scenes, providing actionable insights for future end-to-end architectures.
\end{itemize}

The remainder of this paper is organized as follows. Sect.~\ref{Sect_Related_Work} reviews the related literature. Sect.~\ref{Sect_Methodology} elaborates on the proposed MITE-Net architecture, followed by the introduction of the SAR-Tiny Datasets in Sect.~\ref{Sect_SAR_Tiny}. Sect.~\ref{Sect_Experiment} presents the implementation details and comprehensive experimental results. Finally, Sect.~\ref{Sect_Conclusion} concludes the study.

\section{Related Work}\label{Sect_Related_Work}
In this section, we review the existing literature across three pivotal domains that intersect with our proposed MITE-Net architecture: the fundamental dilemmas in vision-based tiny object detection, the emergence of bio-inspired small target motion detection mechanisms, and the recent advancements in resource-efficient edge perception.

\subsection{Vision-Based Tiny Object Detection}
While deep learning detectors, particularly the YOLO series \cite{redmon2016you, jocher2023ultralytics, jocher2024yolo11}, dominate general visual perception tasks, they face a critical spatial-computational dilemma in 4K imagery where targets typically occupy fewer than 64 pixels. Standard pipelines resize inputs to lower dimensions (e.g., $640 \times 640$) for inference speed, obliterating microscopic features. Conversely, slice-aided strategies like SAHI \cite{akyon2022slicing} preserve fine-grained details but dramatically inflate computational overhead. Recently, alternative paradigms have emerged. Transformer architectures like OWRT-DETR \cite{ma2025owrt} enhance feature representation for maritime UAV search and rescue, while anchor-free models like L-CenterNet \cite{shang2025centernet} mitigate complex anchor-matching in aerial scenarios. Nevertheless, whether relying on global attention or dense convolutions, these models fundamentally operate on a full-frame processing paradigm that remains computationally prohibitive for edge devices.

To circumvent the need for exhaustive slicing, various specialized architectural and training strategies have been developed. Architecturally, Region Proposal Networks (RPNs) \cite{ren2015faster, ren2016faster} generate sparse candidates to filter background noise early, Dilated Backbones \cite{yu2017dilated} maintain high-resolution receptive fields, and CoordConv \cite{liu2018intriguing} explicitly injects spatial awareness. During training, the extreme spatial imbalance necessitates robust optimization. Techniques such as Ground Truth (GT) Jittering \cite{jiang2018acquisition} and Hard Negative Mining \cite{liu2016ssd} ensure robust proposal generation and suppress overwhelming false positives. Furthermore, because standard Intersection over Union (IoU) is highly sensitive to minor pixel shifts at extreme scales, Normalized Wasserstein Distance (NWD) \cite{wang2021normalized} has become a preferred metric for label assignment, typically coupled with Distribution Focal Loss (DFL) \cite{li2020generalized} for precise sub-64-pixel boundary regression.

While our model adopts these advanced spatial and optimization strategies within its detection head to enhance microscopic sensitivity, deploying such dense processing globally across 4K frames remains computationally prohibitive. This motivates our decoupled, cascaded approach. By exploiting bio-inspired motion dynamics for ultra-efficient background filtering, we ensure that computationally intensive, appearance-based recognition is applied exclusively to motion-salient regions, effectively satisfying edge-device constraints.

\subsection{Bio-inspired Small Target Motion Detection}

To circumvent the prohibitive computational burden of dense feature extraction, bio-inspired vision systems offer a promising alternative. In Nature, dragonflies and hoverflies' visual systems efficiently detect minuscule moving objects in complex clutter \cite{Nordstrom2006small, Nordstrom2006insect, Barnett2007retinotopic, Nordstrom2012neural}, which inspires numerous models such as ESTMD \cite{Wiederman2008ESTMD}, DSTMD \cite{Wang2020DSTMD}. Among these, one of the state-of-the-art STMD-based models is the vSTMD \cite{xu2025vstmd}, which introduces cross-inhibition dynamic potential to mine motion cues. These learning-free models excel at suppressing static backgrounds and isolating anomalies with negligible energy consumption, making them ideal for SWaP-constrained platforms.

However, standalone STMDs \cite{Wiederman2008ESTMD, Wang2020DSTMD, Wang2021time, Wang2020STMDpuls, Wang2022attention, ling2022mathematical, Xu2023frac, wang2024BioInspiredSmall, chen2024unveiling, xu2025vstmd} face severe limitations in embodied SAR applications. First, they fundamentally lack high-level semantic representation. While able to detect moving pixels, they cannot classify them, struggling to distinguish critical SAR targets from dynamic environmental noise (e.g., birds or maritime sun glint). Second, these bionic models output discrete coordinates or heatmaps instead of precise bounding boxes, lacking the spatial regression required for downstream UAV tracking.

Consequently, STMDs provide ultra-fast background suppression but they are insufficient as standalone detectors. In this paper, we employ vSTMD as a Region Proposal Network (RPN) and coupled it with a learnable semantic head to bridge the gap between pure motion detection and category-aware bounding box perception.

\subsection{Resource-Efficient Edge Perception}

While traditional UAV-based SAR missions often offload complex processing to ground stations \cite{liu2019uav}, the necessity for immediate, closed-loop responsiveness has driven a paradigm shift toward deploying resource-efficient AI directly onboard aerial platforms \cite{sai2023comprehensive}. However, executing high-resolution (e.g., 4K) perception locally on edge devices introduces severe computational bottlenecks. 

To meet strict Size, Weight, and Power (SWaP) constraints, various lightweight architectures (e.g., MobileNet, YOLO-Nano) and model compression techniques are widely utilized \cite{deng2020model}. Nevertheless, pushing model compression to extreme limits (e.g., sub-1M parameters) inevitably causes representational collapse in tiny target detection. Ultra-lightweight networks simply lack the representational capacity to simultaneously suppress massive background noise and extract the fragile features of sub-64-pixel targets, yielding unacceptably low recall rates in dense aerial imagery.

This exposes a critical gap: the inability of conventional pipelines to decouple background filtering from semantic perception under extreme SWaP budgets. To address this, our proposed MITE-Net leverages the bionic TTM-RPN to explicitly filter irrelevant spatial-temporal backgrounds prior to deep learning inference. This architectural decoupling enables the deployment of an unprecedentedly compact sub-0.14M-parameter head strictly for sparse RoI classification and localization. Ultimately, this approach prevents representational collapse and minimizes inference power to just 3.19 W, providing a highly autonomous and energy-efficient perception paradigm for edge-deployed UAVs.

\section{Proposed Architecture}\label{Sect_Methodology}

\begin{figure*}[!t]
	\centering
	\includegraphics[width=1\textwidth]{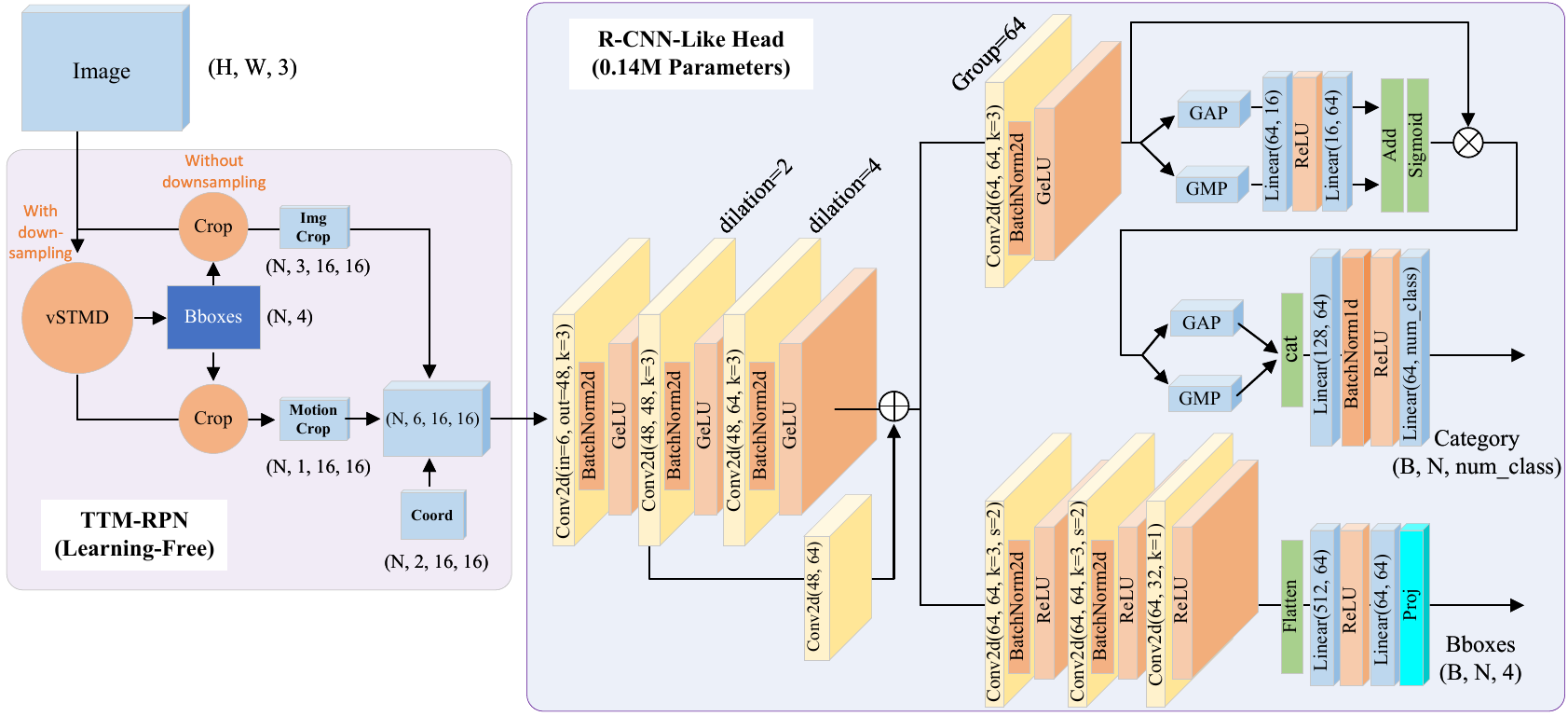}
	\caption{Overall architecture of the proposed MITE-Net. The cascaded framework consists of two stages. Left: A learning-free Bionic-Driven TTM-RPN utilizes vSTMD \cite{xu2025vstmd} to extract motion-salient RoIs, which are cropped and concatenated with spatial coordinates into multi-channel inputs. Right: An ultra-lightweight R-CNN-like Head (<0.14M parameters) processes these inputs using dilated and group convolutions, followed by a dual-branch channel attention structure for precise classification and bounding box regression.}
	\label{Fig_Architecture}
\end{figure*}

In this paper, we propose a cascaded architecture MITE-Net (Motion-Informed Tiny-target Edge Network), as shown in Fig.~\ref{Fig_Architecture}. MITE-Net comprises a learning-free bionic-driven tiny target motion region proposal network (TTM-RPN) and a fully-tailored lightweight learning R-CNN-like head. The TTM-RPN, based on vSTMD \cite{xu2025vstmd}, rapidly extracts regions of interest (RoIs) containing salient motion, which are then cropped and fed into the lightweight head, for precise classification and bounding box regression.

\subsection{Tiny Target Motion-based Region Proposal Network (TTM-RPN)}

The first stage of our pipeline leverages the vSTMD \cite{xu2025vstmd} to generate initial bounding box proposals. Unlike computationally heavy deep learning RPNs \cite{ren2015faster, ren2016faster}, our TTM-RPN is learning-free and lightweight as it relies only on a biologically inspired Spatio-Temporal filtering mechanism.

\subsubsection{TTM Salient Map and Direction}
Operating on a continuous sequence of input frames, vSTMD \cite{xu2025vstmd} first generates a score map and a corresponding direction map for the current observation. The score map quantifies the confidence of tiny-target-induced motion at each spatial pixel, while the direction map encodes its estimated angular trajectory. To suppress background clutter and isolate localized motion, a spatial inhibition layer modeled after center-surround receptive fields (Difference of Gaussians) is applied to the score map. In our approach, we modify this suppression kernel to be size-selective, ensuring robust filtration of specific target dimensions across various downsampling scales (pooling parameters), as detailed in Appendix~\ref{App_Size}. Finally, Non-Maximum Suppression (NMS) and a Top-N selection mechanism are employed to extract the most salient motion points.

\subsubsection{TTM Region Proposal}
Subsequently, we transform these isolated motion points into spatial bounding box proposals. Specifically, we first select the top-$N$ motion candidates ranked by their confidence responses. For each selected candidate, we extract its spatial coordinates $(x, y)$, confidence score, and motion direction $\theta$. Because a detected motion point typically corresponds to the trailing edge of a moving target, we apply a directional forward shift to accurately align the bounding box center with the target body. Given a predefined bounding box size $S$ (e.g., $S=16$), the shift distance $\Delta$ is defined as $0.3 \times S$. Consequently, the adjusted center coordinates $(c_x, c_y)$ of the proposal are calculated as follows:
\begin{subequations}
	\begin{align}
		c_x = x + \Delta \cos(\theta)\\
		c_y = y - \Delta \sin(\theta).  
	\end{align}
\end{subequations}
The final bounding box is defined by the coordinates $[c_x - R, c_y - R, c_x + R, c_y + R]$, with a radius of $R = 0.5 \times S$. We then apply RoIAlign \cite{he2017mask} to map these proposals into fixed 16 × 16 pixel targets.

\subsubsection{TTM Motion-Enhanced Crop}
To provide rich contextual information, we crop not only the raw image channels but also the temporal change and motion response maps from the vSTMD. Let the full-frame feature map be denoted as $F_{combined} = [I_{rgb}, L, V_{resp}]$, where $I_{rgb}$ represents the raw image channels, $L$ is the temporal change layer (encoding signed motion polarity), and $V_{resp}$ is the raw motion response map.

After applying RoIAlign \cite{he2017mask} to the spatial bounding box proposals, we obtain the corresponding local crops: $I_{crop}$, $L_{crop}$, and $V_{crop}$. To effectively utilize the temporal dynamics, the temporal change crop is explicitly separated into positive (ON) and negative (OFF) motion components:
\begin{subequations}
	\begin{align}
		L_{ON} = \max(L_{crop}, 0)\\
		L_{OFF} = \max(-L_{crop}, 0).  
	\end{align}
\end{subequations}
To ensure numerical stability and consistent feature scaling across diverse dynamic scenes, the response and temporal crops are individually normalized by their infinity norm (maximum absolute value).

Finally, we construct an enhanced motion response map, $E_{resp}$, by selectively amplifying the temporal features based on the primary polarity of the detected motion point. Given the polarity mask $p \in \{0, 1\}$ associated with the bounding box proposal (where $p=1$ signifies an ON-polarity motion event and $p=0$ signifies an OFF-polarity event), the enhanced response is formulated as a linear combination:
\begin{align}
	E_{resp} = V_{crop} + p \cdot L_{ON} + (1 - p) \cdot L_{OFF}
\end{align}
The final multi-modal feature tensor, which serves as the input to the subsequent lightweight recognition network, is formed by concatenating the raw image crops and the enhanced response along the channel dimension:
$$T_{input} = [I_{crop}, E_{resp}]$$
This ensures the recognition network simultaneously perceives high-resolution spatial appearances and polarity-aware motion cues.

\subsubsection{Down-Sampling Motion Perception and Raw-Size Crop}
As pixel-wise biological dynamics at native $4\text{K}$ resolution consumes significant power, we introduce a configurable spatial down-sampling mechanism.

In this architecture, the raw input frames ($I_{rgb}$) and their immediate temporal difference ($L$) are natively extracted at the full $4\text{K}$ resolution. When the power-saving mode is activated, an average-pooling factor $ds$ (e.g., $ds=2$) is applied exclusively to the temporal difference before it enters the Leaky Integrate-and-Fire (LIF) pathways (see \cite{xu2025vstmd}):
$$L_{down} = \text{AvgPool2d}(L, \text{kernel\_size}=ds, \text{stride}=ds)$$
Consequently, the computationally intensive membrane potentials ($V_{ON}$ and $V_{OFF}$), as well as the resulting score and direction maps, are computed at a significantly reduced spatial resolution (e.g., $1920 \times 1080$).

To construct the multi-modal feature tensor, bounding box proposals must be accurately aligned with high-frequency 4K visual features. Rather than up-sampling the motion maps, we perform RoIAlign directly in their respective native coordinate spaces: the raw image $I_{rgb}$ is cropped using restored 4K bounding boxes, while the down-sampled maps ($L_{down}$, $V_{down}$) are cropped by setting \textit{spatial\_scale} = $1/ds$. Cropping each modality natively ensures perfect registration into the final $16 \times 16$ tensor.

Crucially, this multi-scale decoupling is applied exclusively during inference. During training, the down-sampling factor is rigidly fixed at $ds=1$ to ensure the lightweight head learns the most accurate, fine-grained spatial-motion correlations, which subsequently allows the UAV to aggressively scale down power consumption during field deployment without compromising learned representations.

\subsection{Lightweight R-CNN-like Head}
To maintain high detection accuracy for microscopic objects while restricting the parameter footprint to under 0.14M, the R-CNN-like Head is specifically tailored to process the 16 × 16 RoI features, a resolution explicitly chosen to align with the typical spatial scale of tiny targets.

\subsubsection{Coordinate Convolution (CoordConv)} 
We explicitly inject normalized spatial coordinates $x$ and $y$ ($[-1, 1]$) into the input tensor, allowing the network to retain absolute spatial awareness within the cropped region \cite{liu2018intriguing}.

\subsubsection{Dilated Backbone}
The feature extractor avoids spatial downsampling (pooling) to preserve the fragile features of tiny targets. Instead, it employs a cascade of standard and dilated convolutions (dilation rates of 2 and 4) with residual connections \cite{yu2017dilated}, rapidly expanding the receptive field to cover the entire $16 \times 16$ patch.

\subsubsection{Decoupled Heads \cite{ge2021yolox}}
The classification head utilizes a spatial mixing Depthwise Convolution \cite{howard2017mobilenets} followed by a dual-attention mechanism (Max and Average pooling) to capture both dominant foreground and contextual background cues. In addition, the regression head utilizes a convolutional bottleneck mapping features to a discrete Distribution Focal Loss (DFL) space \cite{li2020generalized}.

\subsection{Training Strategy}
Training a cascaded detector for tiny targets presents unique challenges, particularly regarding sample imbalance and label assignment.

\subsubsection{GT Jittering for Robust Proposal Generation}
Since vSTMD is learning-free, we simulate realistic, imperfect motion proposals during training by applying random spatial jittering to the Ground Truth (GT) bounding boxes. We apply scale perturbations of $\pm 20\%$ and spatial shifts to the GT boxes, effectively generating a rich distribution of positive RoI crops \cite{jiang2018acquisition}.

\subsubsection{NWD-Based Label Assignment}
For robust label assignment, we bypass the classic Intersection over Union (IoU) in favor of the Normalized Wasserstein Distance (NWD) \cite{wang2021normalized}, a validated metric explicitly suited for tiny targets. By modeling the predicted box $p$ and the target box $t$ as 2D Gaussian distributions, the NWD distance $d_{nwd}$ is calculated as follows:
\begin{align}
	d_{nwd} = \exp\bigg( - \frac{1}{C} * \Big[ & (cx_p - cx_t)^2 + (cy_p - cy_t)^2 \nonumber \\
	&+ \frac{(w_p - w_t)^2 + (h_p - h_t)^2}{4}\Big]^{\frac{1}{2}} \bigg)
	\label{Eq_nwd}
\end{align}
where $C$ is a dataset-specific normalization constant (set to $12.8$). We calculate the NWD similarity between the motion proposals and the ground-truth boxes. Proposals exceeding an NWD threshold (typically $0.3 \sim 0.5$) are designated as positive samples.

\subsubsection{Hard Negative Mining}
To combat the overwhelming number of background proposals, we employ a strict 1:3 positive-to-negative sampling ratio \cite{liu2016ssd}. We select the highest-scoring negative samples based on the classification logits to participate in the loss calculation, effectively penalizing false alarms generated by the motion detector.

\begin{table*}[!t]
	\centering
	\caption{Detailed Sequence Allocation and Attributes of the 4K SAR-Tiny Datasets for Embodied Edge SAR}
	\label{Tab_SAR_Tiny_Datasets}
	\begin{tabular}{ccccccp{5.5cm}}
		\toprule
		\textbf{Dataset} & \textbf{Split} & \textbf{Sequences} & \textbf{Frames} & \textbf{BBoxes} & \textbf{Target Density} & \textbf{Key Characteristics} \\ 
		\midrule
		\multirow{3}{*}[-2ex]{\textbf{SeaDroneSee-Tiny}} 
		& Train & seq~2, 3, 4, 5, 6, 7, 8 & 3858 & 7178 & 1-5 / frame & Dynamic maritime backgrounds. \\ \cmidrule(l){2-7} 
		& Val   & seq~9    & 1001 & 3003 & 3 / frame  & Baseline maritime background evaluation. \\ \cmidrule(l){2-7} 
		& Test  & seq~1       & 1001 & 11243& 12 / frame  & Density mimicking real-world SAR crises. \\ \midrule
		\multirow{3}{*}[-2ex]{\textbf{UAVID-Tiny}}       
		& Train & seq~2, 5, 7, 8, 16, 17, 33 & 3208 & 63398 & 10-50/ frame & Massive urban clutter. \\ \cmidrule(l){2-7} 
		& Val   & seq~24         & 901   & 15063 & $\sim$16 / frame & Occlusion and varied lighting conditions. \\ \cmidrule(l){2-7} 
		& Test  & seq~23        & 701   & 9839 & $\sim$14 / frame & Early-stage target discovery in urban scenes. \\ 
		\bottomrule
	\end{tabular}
\end{table*}

\subsection{Loss Function Criteria}

The overall loss function for R-CNN-like Head is a weighted sum of classification, regression, and distribution losses:
\begin{align}
L_{total} = \lambda_{cls} L_{cls} + \lambda_{reg} L_{nwd} + \lambda_{dfl} L_{dfl}
\end{align}
where the empirical weights are set to $\lambda_{cls} = 1.0$, $\lambda_{reg} = 5.0$, and $\lambda_{dfl} = 0.5$.

\subsubsection{Classification Loss}
We utilize Binary Cross-Entropy (BCE) \cite{lecun2015deep} with logits over the sampled subset $S$ (containing the 1:3 pos/neg ratio), normalized by the number of positive samples $N_{pos}$:
\begin{align}
	L_{cls} = \frac{1}{N_{pos}} \sum_{i \in S} BCE(p_i, y_{i}^{cls})
\end{align}

\subsubsection{Bounding Box Regression Loss}
Consistent with our label assignment strategy, we directly optimize the Normalized Wasserstein Distance (NWD) \cite{wang2021normalized} for robust bounding box regression. NWD loss are calculated as follows:
\begin{align}
	L_{nwd} =  1 - d_{nwd}
\end{align}

\subsubsection{Distribution Focal Loss}
To handle the uncertainty and ambiguity of tiny target boundaries, we regress the distances from the center to the four box edges using Distribution Focal Loss (DFL) \cite{li2020generalized}. For a given edge, let $d \in \mathbb{R}$ represent the continuous ground-truth distance. We bound $d$ by its two adjacent discrete bins, $d_i = \lfloor d \rfloor$ and $d_{i+1} = \lfloor d \rfloor + 1$. The network predicts a probability distribution over these discrete bins, and the loss interpolates the cross-entropy between the two nearest bins:
\begin{align}
	L_{dfl} = - \sum_{e \in \{l, t, r, b\}} &\left[ (d_{i+1}^{(e)} - d^{(e)}) \log(S_i^{(e)}) \right.\nonumber 
	\\ & \quad + \left. (d^{(e)} - d_i^{(e)}) \log(S_{i+1}^{(e)}) \right]
\end{align}
where $S_i^{(e)}$ and $S_{i+1}^{(e)}$ represent the Softmax probabilities of the regression logits at bins $i$ and $i+1$ for edge $e$ (left, top, right, bottom). The integration bounds are set to $reg\_max = 16$, perfectly corresponding to the maximum pixel width of the RoI crop.

\begin{figure*}[!t]
	\centering
	\includegraphics[width=1\textwidth]{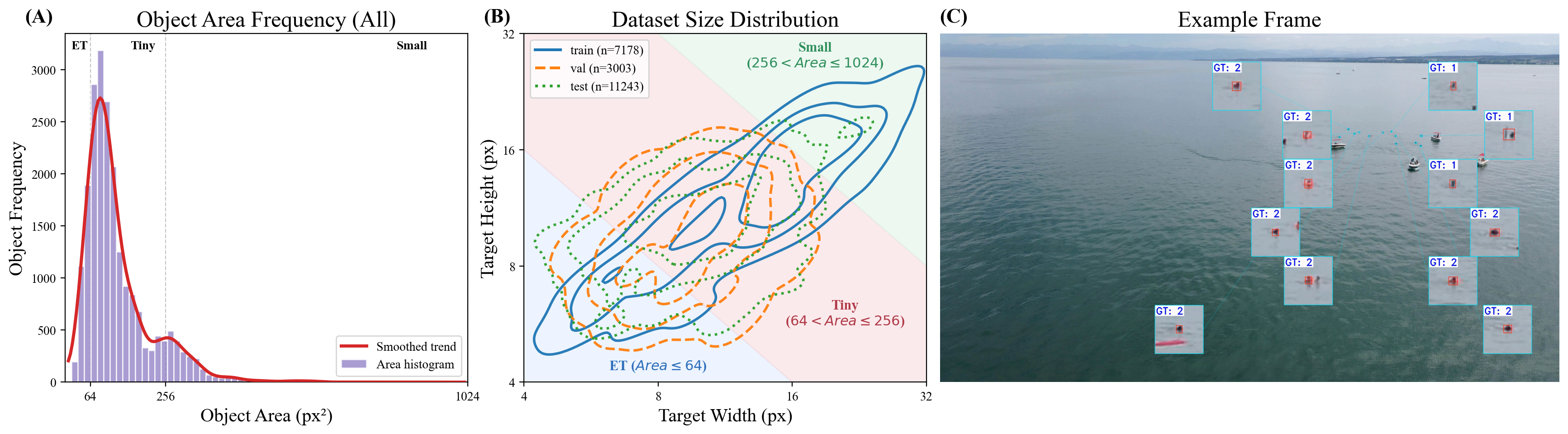}
	\caption{Statistical overview and visual example of the curated SeaDroneSee-Tiny dataset. (A) Object area frequency distribution across the entire dataset, demonstrating a heavily right-skewed trend, where the majority of targets are categorized as Extremely Tiny (ET, Area $\le$ 64 px$^2$) and Tiny (64 $<$ Area $\le$ 256 px$^2$). (B) Joint distribution of target width and height across the training, validation, and test splits, partitioned by the three defined scale regimes: ET, Tiny, and Small (256 $<$ Area $\le$ 1024 px$^2$). The highly concentrated contour of the test set in the ET and Tiny regions highlights the deliberate "stress test" design. (C) An example frame from the dataset illustrating the visual challenges of detecting multiple scattered targets. The bounding boxes denote ground truth (GT) labels inherited from the original dataset: GT 1 represents a swimmer, and GT 2 represents a swimmer with a life jacket.}
	\label{Fig_SeaDroneSee_Tiny}
\end{figure*}

\section{4K SAR-Tiny Datasets}\label{Sect_SAR_Tiny}

To rigorously evaluate embodied edge Search and Rescue (SAR) missions, high-quality, high-resolution datasets with a strict focus on tiny targets are required. Existing Unmanned Aerial Vehicle (UAV) datasets often lack the 4K resolution necessary for early-stage target discovery or do not provide dedicated annotations for highly degraded, minute targets. To bridge this gap, we constructed the SAR-Tiny Datasets by extracting and relabeling two distinct subsets from publicly available 4K UAV datasets: SeaDroneSee-MOT~\cite{kiefer20221st, varga2022seadronessee} and UAVID~\cite{lyu2020uavid}. All relabeling procedures were conducted using CVAT\footnote{\url{https://www.cvat.ai/}} \cite{cvat}, guided by a rigorous annotation protocol and a multi-stage quality assurance pipeline, which are detailed in Appendix~\ref{App_CVAT}.

We specifically filtered these datasets to focus on minute bounding boxes, categorizing them based on pixel area: \textbf{Extremely Tiny} (ET, Area $\le$ 64 pixels), \textbf{Tiny} (64 pixels $<$ Area $\le$ 256 pixels), and \textbf{Small} (256 pixels $<$ Area $\le$ 1024 pixels), following \cite{ying2025VisibleThermalTiny}. 

These datasets offer escalating difficulty: SeaDroneSee-Tiny features dynamic maritime environments with predominantly tiny targets, whereas UAVID-Tiny introduces highly cluttered urban landscapes where severely degraded targets shift heavily into the extremely tiny (ET) domain. To avoid data leakage and background memorization, both datasets are split by video sequences: 7 for training, 1 for validation, and 1 for testing.

\subsection{SeaDroneSee-Tiny}

The SeaDroneSee-MOT dataset \cite{kiefer20221st, varga2022seadronessee} was originally designed for Multiple Object Tracking in maritime environments. Based on it, we curated the SeaDroneSee-Tiny dataset by selectively extracting and entirely relabeling frames containing minute targets across dynamic maritime scenes, where we retained the human-centric categories from the original annotations, including swimmers (Class 1) and swimmers with life jackets (Class 2). The `boat' category was deliberately discarded, as these instances predominantly consisted of large targets that do not align with the strict scale constraints of our study.

As detailed in Table~\ref{Tab_SAR_Tiny_Datasets}, the SeaDroneSee-Tiny dataset is systematically partitioned to evaluate model robustness under varying target distributions and complex maritime backgrounds. Specifically, the training set aggregates sequences 2 through 8, providing 3858 frames with sparse target distributions (1 to 5 targets per frame) to learn generalizable features under dynamic ocean surface conditions. The validation set (seq~9) serves as a baseline for performance evaluation. The test split is exclusively designated to seq~1, which mimics real-world crises by featuring an exceptionally simultaneous presence of 12 minuscule targets per frame.

The target size distribution is illustrated in Fig.~\ref{Fig_SeaDroneSee_Tiny}(A) and the unusually concentrated test contour in Fig.~\ref{Fig_SeaDroneSee_Tiny}(B). While the training set contains 7178 annotations, the test set intentionally comprises 11243 annotations. Based on our tri-level scale (ET, Tiny, Small) classification, the smoothed trend exhibits a sharp peak around the boundary of the ET and Tiny regions. The vast majority of target areas fall strictly below the 256 px$^2$ threshold. This closely mimics real-world maritime SAR emergencies where an edge-deployed UAV must suddenly detect numerous scattered, minuscule targets amidst complex ocean backgrounds.

\subsection{UAVID-Tiny}
\begin{figure*}[!t]
	\centering
	\includegraphics[width=1\textwidth]{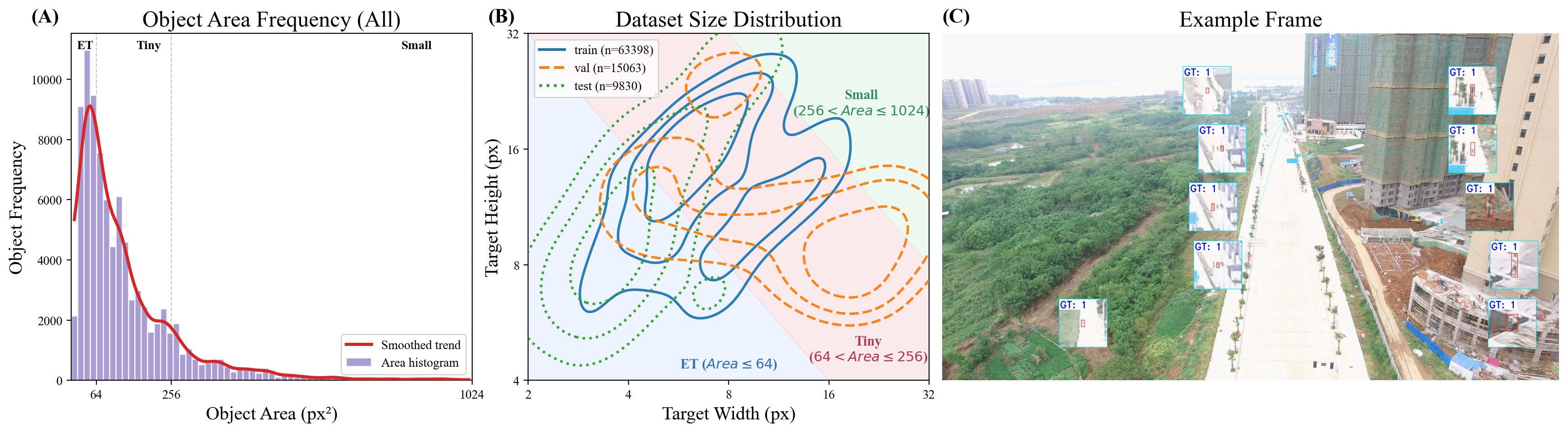}
	\caption{Statistical overview and visual example of the curated UAVID-Tiny dataset. (A) Object area frequency distribution, highlighting a massive concentration of targets strictly within the Extremely Tiny (ET, Area $\le$ 64 px$^2$) region. (B) Joint distribution of target width and height across data splits, where the deep penetration into the ET region underscores the dataset's complexity. (C) Example test frame of minute targets in a cluttered urban background. The bounding boxes denote ground truth (GT) labels: GT 1 represents pedestrians, while the dataset also includes moving (GT 2) and static cars (GT 3).}
	\label{Fig_UAVID_Tiny}
\end{figure*}

The original UAVID dataset \cite{lyu2020uavid} is a high-resolution UAV video dataset introduced for semantic segmentation. Among them, we extracted 9 sequences including ET-targets and performed meticulous bounding box re-annotation to create the UAVID-Tiny dataset. During this process, we shifted the primary focus to humans (Class 1) while inheriting the moving car (Class 2) and static car (Class 3) categories. Background categories originally used for instance segmentation, such as buildings and roads, were discarded.

As detailed in Table~\ref{Tab_SAR_Tiny_Datasets}, the UAVID-Tiny dataset is partitioned to rigorously evaluate model generalization in complex urban environments. The training set (sequences 2, 5, 7, 8, 16, 17, 33) provides a massive corpus of 63,398 annotations across 3,208 frames, forcing the model to learn robust features against severe urban clutter. The validation set (sequence 24) introduces 15,063 bounding boxes to test performance under varying lighting and occlusions. Finally, the test set (sequence 23) comprises 9,839 annotations across 701 frames, specifically challenging the model's capacity for early-stage target discovery in visually deceptive scenes.

The impact of this urban curation is visualized in Fig.~\ref{Fig_UAVID_Tiny}(A). Based on our tri-level scale classification, the object area histogram exhibits a sharp concentration strictly within the ET domain, peaking near the 64 px$^2$ threshold. Furthermore, the KDE contour plot in Fig.~\ref{Fig_UAVID_Tiny}(B) demonstrates that target dimensions penetrate deeply into the ET and Tiny regions, heavily skewed toward sub-64 pixel areas. This combination of scattered distribution, extreme scale limitations, and severe background clutter establishes a rigorous testing ground for embodied edge SAR perception beyond maritime environments.

\section{Implementation and Experiment}\label{Sect_Experiment}
In this section, we first outline the experiment settings and Implementation details, followed by the mission-level benchmark results and a detailed system analysis. Additional supplementary experiments are provided in the appendices.

\subsection{Experiment Settings}
\subsubsection{Compared Models} \label{Sect_Compared_Model}
To ensure a fair and practical evaluation for embodied edge intelligence, the selected baselines must satisfy two strict criteria: (1) native support for edge-device compilation (e.g., TensorRT/ONNX export compatibility) without relying on complex, unoptimized custom operators, and (2) the capability to achieve real-time inference ($\geq 30$ FPS) on the NVIDIA Jetson platform. Consequently, classical two-stage detectors (e.g., standard Faster R-CNN~\cite{ren2015faster, ren2016faster}), traditional motion-estimation pipelines (e.g., dense optical flow~\cite{alfarano2024estimating}), and Transformer-based architectures (e.g., DETR variants~\cite{ma2025owrt}), which suffer from severe latency bottlenecks or lack mature edge-deployment toolchains, are explicitly excluded from hardware-level benchmarking. Accordingly, we benchmark the proposed MITE-Net against three representative state-of-the-art (SOTA) configurations:
\begin{itemize}
	\item \textbf{YOLOv11n} \cite{jocher2024yolo11}: A highly representative milestone within the YOLO family, representing the cutting edge of lightweight, real-time object detection. It offers an exceptional balance of accuracy and computational efficiency, making it suitable for edge devices.
	\item \textbf{YOLOv8n-P2 (Standard Resize)} \cite{jocher2023ultralytics}: A specialized variant of YOLOv8 explicitly tailored for small object detection by incorporating high-resolution P2 feature maps. In this configuration, the original 4K video frames are directly downsampled to the target network input size.
	\item \textbf{YOLOv8n-P2 (SAHI)} \cite{akyon2022slicing}: Utilizing the identical P2-enhanced architecture, this setup applies Slicing Aided Hyper Inference (SAHI) on the native 4K images. This approach processes high-resolution patches independently, mitigating the spatial information loss caused by standard resizing.
\end{itemize}

\subsubsection{Evaluation Criteria}
Unlike conventional vision tasks that rely primarily on static bounding box overlap, evaluating an embodied edge SAR system requires a multidimensional approach. We comprehensively validate our model across two principal domains:

\textbf{SAR Task Effectiveness:} Prioritizing the reliable discovery of life-critical targets:
\begin{itemize}
	\item \textbf{Search Success Rate:} The proportion of GT trajectories locked under an $m$-out-of-$k$ criterion. Frame-level associations are determined via one-to-one Hungarian matching (using NWD cost) to strictly resolve overlapping detections. A GT is successfully locked if matched (score $> 0.5$, $d_{nwd}> 0.5$) in $\ge m$ frames within a $k$-frame sliding window, where $m=3$ and $k=5$ in this paper.
	\item \textbf{False Alarm Rate:} The ratio of incorrect detections to total frames. While secondary to success rate, minimizing false alarms is vital to prevent unnecessary battery depletion on invalid targets.
	\item \textbf{Search Time:} The average frames elapsed from a target's initial appearance until the $m$-out-of-$k$ criterion is met. Lower time-to-search reflect superior early-stage, long-range perception.
\end{itemize}

\textbf{SWaP and Hardware Efficiency:} To validate deployment feasibility on edge devices, we profile:
\begin{itemize}
	\item \textbf{Model Scale:} Total trainable parameters (M), computational complexity (G-FLOPs), and GPU memory footprint (GB).
	\item \textbf{Power Consumption:} Average overall system power and inference-specific power (Watts), where inference power encompasses only model inference and pre/post-processing, excluding image reading.
	\item \textbf{Hardware Efficiency:} Inference latency (ms, batch=1) and energy efficiency (FPS/Watt), which directly dictates the UAV's operational flight endurance.
\end{itemize}

\begin{table*}[t]
	\caption{Comprehensive Comparison of Search and Rescue (SAR) and Size, Weight, and Power (SWaP) Metrics on the SeaDroneSee-Tiny (SDS-T) and UAVID-Tiny (U-T) Datasets. Inference was conducted on an NVIDIA Jetson AGX Xavier using Float16 precision. Highlighted rows indicate primary evaluation criteria. Bold and underlined values represent the optimal and sub-optimal performances, respectively. Grayed text denotes severe algorithmic degradation or failure.}
	\label{Tab_Comparison}
	\centering
	\begin{tabular}{c c r c c c c c}
		\toprule
		&&& YOLOv11n \cite{jocher2024yolo11} & \multicolumn{2}{c}{YOLOv8n-P2 \cite{jocher2023ultralytics}} &  \multicolumn{2}{c}{MITE-Net (Proposed)} \\
		\cmidrule(lr){4-4} \cmidrule(lr){5-6} \cmidrule(lr){7-8}
		&&& \parbox{2cm}{\centering Resize\\$1920\times1088$} & \parbox{2cm}{\centering Resize\\$1920\times1088$} & \parbox{2cm}{\centering SAHI\cite{akyon2022slicing}\\$1344\times768\times9$} & \parbox{2cm}{\centering \textbf{Raw}\\\textbf{4K + 2ds}} & \parbox{2cm}{\centering Raw\\4K + 8ds} \\
		\midrule
		\multirow{8}{*}{\rotatebox{90}{\parbox{1cm}{\centering SAR\\Criteria}}}
		&\multirow{4}{*}{SDS-T}
		&\multicolumn{1}{>{\columncolor{yellow!20}}r}{\textbf{Search Success Rate} (\%, $\uparrow$)} 
		& \multicolumn{1}{>{\columncolor{yellow!20}}c}{83.33}  &   \multicolumn{1}{>{\columncolor{yellow!20}}c}{83.33}   &     \multicolumn{1}{>{\columncolor{yellow!20}}c}{\textbf{100}}     &    \multicolumn{1}{>{\columncolor{yellow!20}}c}{\textbf{100}}   &    \multicolumn{1}{>{\columncolor{yellow!20}}c}{83.33} \\
		&&False Alarm Rate (\%, $\downarrow$)  
		& 6.83  &   16.77   &   3.12   &   67.72   & 82.29  \\
		&&Max Search Time (Frame, $\downarrow$) 
		& 126 &   74   &     385     &     299    &  119    \\
		&&Avg. Search Time (Frame, $\downarrow$) & 17.90  &   11.20   &     62.58      &     69.00     &  36.80    \\
		\cmidrule(lr){2-3}
		& \multirow{4}{*}{U-T} 
		&\multicolumn{1}{>{\columncolor{yellow!20}}r}{\textbf{Search Success Rate} (\%, $\uparrow$)} 
		& \multicolumn{1}{>{\columncolor{yellow!20}}c}{11.11}  
		& \multicolumn{1}{>{\columncolor{yellow!20}}c}{\underline{22.22}}  
		& \multicolumn{1}{>{\columncolor{yellow!20}}c}{\textbf{36.11}}     
		& \multicolumn{1}{>{\columncolor{yellow!20}}c}{{\color{gray}8.33}} 
		& \multicolumn{1}{>{\columncolor{yellow!20}}c}{{\color{gray} 0.00}} \\
		&&False Alarm Rate (\%, $\downarrow$)  & 36.90  & 54.87 & 21.35 & {\color{gray}99.39} & {\color{gray} 99.98} \\
		&&Max Search Time (Frame, $\downarrow$) & 32 & 696 & 512 & 353.00  & -  \\
		&&Avg. Search Time (Frame, $\downarrow$) & 21.00 & 99.88 & 146.77 & 299.67 &  -  \\
		\cmidrule(lr){1-3}
		\multirow{8}{*}{\rotatebox{90}{\parbox{2cm}{\centering SWaP\\Criteria}}}&&Parameters (M, $\downarrow$) 
		& \underline{2.59} & 3.01 & 2.93 & \textbf{0.14} & \textbf{0.14}  \\
		&&Complexity (G-FLOPs, $\downarrow$) 
		& 32.85 & 41.80 & 31.15 & \underline{0.56} & \textbf{0.53} \\
		&&GPU Memory (GB, $\downarrow$) 
		& \underline{2.46} &  2.55  &    2.49     &    2.56    &  \textbf{2.11}   \\
		& &\multicolumn{1}{>{\columncolor{yellow!20}}r}{\textbf{Latency} (ms, Batch=1, $\downarrow$)} 
		& \multicolumn{1}{>{\columncolor{yellow!20}}c}{34.31}  &   \multicolumn{1}{>{\columncolor{yellow!20}}c}{\underline{32.03}}   &   \multicolumn{1}{>{\columncolor{yellow!20}}c}{\color{gray}{208.68}}     &     \multicolumn{1}{>{\columncolor{yellow!20}}c}{32.97}     &     \multicolumn{1}{>{\columncolor{yellow!20}}c}{\textbf{19.11}} \\
		&&Total Power (Watt, $\downarrow$)  
		& 27.56  &   27.99   &     29.39      &     \underline{21.10}     & \textbf{18.94}      \\
		&&Inference Power (Watt, $\downarrow$)  
		& 9.91  &   10.15   &     13.53      &     \underline{3.19}    & \textbf{1.41}      \\
		&&\multicolumn{1}{>{\columncolor{yellow!20}}r}{\textbf{Efficiency} (FPS/Watt, $\uparrow$) } 
		& \multicolumn{1}{>{\columncolor{yellow!20}}c}{2.94} & \multicolumn{1}{>{\columncolor{yellow!20}}c}{3.08} & \multicolumn{1}{>{\columncolor{yellow!20}}c}{0.35} & \multicolumn{1}{>{\columncolor{yellow!20}}c}{\underline{9.51}} & \multicolumn{1}{>{\columncolor{yellow!20}}c}{\textbf{13.54}}  \\
		\bottomrule
	\end{tabular}
\end{table*}

\subsubsection{Datasets}
The proposed method is evaluated on our SeaDroneSee-Tiny and UAVID-Tiny datasets, see in Sect~\ref{Sect_SAR_Tiny}.

\subsection{Implementation Details}
To rigorously evaluate both perception accuracy and SWaP efficiency, our experiments are conducted across two distinct hardware platforms. Model training and validation are performed on a workstation equipped with an NVIDIA GeForce RTX 5060 Ti GPU (16GB VRAM), running Python 3.12 and CUDA 13.0. Real-world edge deployment and power profiling are executed on an NVIDIA Jetson AGX Xavier (32GB shared memory). The edge software environment is built on JetPack 5.1.4, Python 3.8.10, and PyTorch 2.1.0.

All comparison YOLO models are post-trained on the proposed SeaDroneSee-Tiny and UAVID-Tiny datasets, with leveraging official Ultralytics 8.4.30 framework. Model checkpoints are selected based on the highest $F_{2}$ score achieved on the validation set. To simulate realistic UAV operations, input dimensions during both training and testing are carefully scaled to the maximum resolution that strictly satisfies real-time processing constraints ($\approx 30$ FPS) while consistently preserving the native 16:9 aspect ratio of 4K videos. 

Finally, to maximize inference efficiency and adhere to strict SWaP constraints on the edge device, all optimal PyTorch models are first exported to the ONNX format and subsequently compiled into highly optimized TensorRT engines using half-precision (FP16) quantization.

\subsection{SAR Mission-Level Benchmark}

We divide the comparisons into two core dimensions: mission-critical search capabilities (SAR Criteria) and hardware-level deployment efficiency (SWaP Criteria). Table~\ref{Tab_Comparison} presents the quantitative results, with representative visualizations in Fig.~\ref{Fig_Output}.

\subsubsection{SAR Performance Evaluation}
On the maritime SeaDroneSee-Tiny (SDS-T) dataset, MITE-Net (2ds) directly processes raw 4K imagery achieving a perfect 100\% Search Success Rate (SSR). Conversely, resized YOLO baselines drop to 83.33\% due to the loss of microscopic features. While YOLOv8n-P2 with SAHI restores SSR to 100\%, it severely violates SWaP constraints. Notably, under aggressive spatial compression (8ds), MITE-Net's SSR drops to 83.33\%, matching the resized YOLO baselines but at a fraction of their computational cost.

Admittedly, MITE-Net yields a higher False Alarm Rate (FAR) of 67.72\% (2ds) and 82.29\% (8ds). However, in life-critical SAR protocols, preventing missed detections strictly supersedes bounding-box precision. Furthermore, this FAR can be effectively managed by tuning the proposal limit $N$, as previously analyzed in Table~\ref{Tab_Scalability}.

Conversely, evaluation on the ultra-cluttered UAVID-Tiny (U-T) dataset reveals the architecture's boundaries in static-dominated urban scenes. While the YOLO-based baselines retain partial effectiveness (SSRs ranging from 11.11\% to 36.11\%), MITE-Net suffers a severe performance collapse, yielding an SSR of 8.33\% (2ds) and total algorithmic failure at 8ds (0.00\%). The specific structural limitations causing this degradation are analyzed in detail in Sect.~\ref{Sect_Module_Analysis}, and comparisons with detection-level criteria are showcased in Appendix~\ref{App_Comparison}.

\begin{table*}[!t]
	\centering
	\caption{Model scalability of the proposed MITE-Net concerning downsampling factor ($ds$) and proposal bounding-box number ($N$) on SeaDroneSee-Tiny. SSR: Search Success Rate (\%); FAR: False Alarm Rate (\%); FPS: Frames Per Second; Eff.: Efficiency (FPS/Watt); FLOPs: Floating Point Operations (G). Up/down arrows indicate the preferred direction of the metric.}
	\label{Tab_Scalability}
	\begin{tabular}{cccccccccc}
		\toprule
		\multicolumn{2}{c}{\textbf{Configurations}} & \multicolumn{2}{c}{\textbf{SAR Criteria}} & \multicolumn{6}{c}{\textbf{SWaP Criteria}} \\
		\cmidrule(lr){1-2} \cmidrule(lr){3-4} \cmidrule(lr){5-10}
		\multirow{2}{*}{\textit{ds}} & \multirow{2}{*}{$N$} & \multirow{2}{*}{SSR ($\uparrow$)} & \multirow{2}{*}{FAR ($\downarrow$)} & \multirow{2}{*}{FPS ($\uparrow$)} & \multirow{2}{*}{Eff. ($\uparrow$)} 
		& \multirow{2}{*}{Total FLOPs (G, $\downarrow$)} & \multicolumn{2}{c}{TTM-RPN FLOPs (G, $\downarrow$)} & \multirow{2}{*}{\parbox{3cm}{\centering R-CNN-like Head\\FLOPs (G, $\downarrow$)}}  \\
		\cmidrule(lr){8-9}
		&&&&&&&vSTMD & Others&\\
		\midrule
		1 & 25  & 100.00 & 67.83 & 11.81 & 7.43 
		&2.1851 & 1.9972 & 0.0317 & 0.1561 \\
		\cmidrule(lr){1-2}
		\multirow{4}{*}{\textbf{2}} & 10  & 100.00 & 35.41 & 30.94 & 10.18 
		&0.5268 & 0.3689 & 0.0258 & 0.1321\\
		& \textbf{25}  & 100.00 & 67.72 & 30.33 & 10.32 
		&0.5605 & 0.3689 & 0.0258 & 0.1657\\
		& 50  & 100.00 & 82.76 & 30.16 & 10.12 
		& 0.5607 & 0.3689 & 0.0258 & 0.1659\\
		& 100 & 100.00 & 82.76 & 30.16 & 10.12 
		& 0.5607 & 0.3689 & 0.0258 & 0.1659\\
		\cmidrule(lr){1-2}
		4 & 25 & 100.00 & 68.80 & 46.28 & 12.63
		& 0.4335 & 0.1290 & 0.0246 & 0.2799 \\
		8 & 25 & 83.33 & 82.29 & 52.33 & 13.54
		& 0.5167 & 0.0925 & 0.0251 & 0.3991 \\
		\bottomrule
	\end{tabular}%
\end{table*}

\subsubsection{SWaP Analysis}
MITE-Net operates with merely 0.14M parameters and sub-0.6 G-FLOPs complexity (0.56 for 2ds, 0.53 for 8ds)—orders of magnitude lower than the YOLO baselines ($\sim$3M parameters, $>30$ G-FLOPs).

This ultra-lightweight footprint resolves the standard latency-accuracy dilemma. While YOLOv8n-P2 with SAHI achieves 100\% SSR, its prohibitive latency (208.68~ms) precludes real-time UAV operations. Resized YOLO models are fast ($\sim$32--34~ms) but sacrifice accuracy. MITE-Net elegantly bridges this gap: the 2ds configuration processes 4K images at a real-time 32.97~ms, while 8ds unlocks an ultra-fast 19.11~ms.

Critically, MITE-Net (2ds) consumes only 3.19~W of inference power, yielding an exceptional energy efficiency of 9.51 FPS/Watt. The 8ds configuration pushes this boundary further to just 1.41~W and 13.54 FPS/Watt. Both settings significantly outperform standard YOLO configurations ($\sim$3.0 FPS/Watt) and the SAHI approach (0.35 FPS/Watt), confirming MITE-Net as an optimally scalable solution for edge-deployed UAVs.

\subsection{System Analysis}
\subsubsection{Scalability and System Trade-offs}

Table~\ref{Tab_Scalability} evaluates MITE-Net's architectural scalability across varying spatial downsampling factors ($ds$) and proposal limits ($N$) on the SeaDroneSee-Tiny dataset, providing a comprehensive assessment of both SAR performance and SWaP constraints.

\textbf{Robust Target Recall:} MITE-Net consistently maintains a 100\% Search Success Rate (SSR) across configurations up to $ds=4$. This demonstrates the TTM-RPN's robustness against representational collapse, even under severe spatial degradation ($ds=4, N=25$) and stringent proposal throttling ($ds=2, N=10$). However, extreme spatial compression at $ds=8$ exceeds the representational limit, resulting in a recall drop to 83.33\%.

\textbf{The Proposal ($N$) Trade-off:} Increasing $N$ beyond 10 yields no marginal recall benefit. Instead, excess proposals force the classification head to process more background noise. This not only sharply inflates the False Alarm Rate (FAR) up to 82.76\% (at $N=50$ and $N=100$), but also predictably increases the R-CNN-like Head computational overhead (rising from 0.1321 to 0.1659 G-FLOPs at $ds=2$), slightly degrading inference speed. We adopt $N=25$ as a conservative optimum to buffer against environmental dynamics while strictly bounding the computational load.

\textbf{SWaP Optimization and Computational Shift via Spatial Scaling ($ds$):} The downsampling factor $ds$ serves as the primary lever for hardware scalability, revealing a notable shift in the system's computational bottleneck. At $ds=1$, the vSTMD module strictly dominates the computational cost (accounting for 1.9972 of the 2.1851 Total G-FLOPs). Scaling to $ds=2$ drastically slashes the Total FLOP by over 74\% (to 0.5605~G), optimally balancing real-time perception (30.33 FPS) with energy efficiency (10.32 FPS/Watt). Aggressively scaling to $ds=4$ further reduces the Total FLOPs to 0.4335~G, unlocking an ultra-efficient mode (46.28 FPS, 12.63 FPS/Watt) without sacrificing target detections. Interestingly, as $ds$ increases, the primary computational burden transitions: at $ds=4$ and $ds=8$, the R-CNN-like Head FLOP overtakes the drastically reduced TTM-RPN FLOP to become the main computational overhead.

\subsubsection{Module Analysis} \label{Sect_Module_Analysis}

\begin{table}[!t]
	\centering
	\caption{Module-wise performance analysis of the proposed two-stage MITE-Net. $R_a$ denotes the recall at an NWD threshold of $a$. The significant degradation on UAVID-Tiny highlights the limitations of the ultra-lightweight head and non-learning RPN in complex urban scenes.}
	\label{Tab_Module}
	\begin{tabular}{ccccc}
		\toprule
		\multirow{2}{*}{Module}  &\multicolumn{2}{c}{SeaDroneSee-Tiny}
		&\multicolumn{2}{c}{UAVID-Tiny} \\
		\cmidrule(lr){2-3} \cmidrule(lr){4-5}
		& R$_{0.1}$ & R$_{0.5}$ &R$_{0.1}$ & R$_{0.5}$  \\
		\midrule
		\parbox{2cm}{\centering TTM-RPN\\ (No-Category)} & 0.9584 & 0.5812 & 0.1047 & 0.0580  \\
		\cmidrule(lr){1-1}
		\parbox{2cm}{\centering R-CNN-Head\\ (Category)} & 0.7377 &  0.7072 & 0.0230 & 0.0103  \\
		\bottomrule
	\end{tabular}%
\end{table}

We conduct a module-wise performance analysis across different scenarios, as summarized in Table \ref{Tab_Module}. While the model demonstrates robust performance on the SeaDroneSee-Tiny dataset, it encounters severe training collapse and performance degradation on the UAVID-Tiny dataset. This failure can be primarily attributed to two structural limitations:

First, the non-learning nature of the Tiny Target Motion RPN (TTM-RPN) limits its robustness in complex scenes. The UAVID-Tiny dataset features highly cluttered urban backgrounds, which severely interfere with motion-based target extraction. Consequently, the TTM-RPN achieves an extremely low initial hit rate of only 0.1047 ($R_{0.1}$), indicating that the current TTM-RPN struggles to reliably separate tiny targets from complex urban noise.

Second, the restricted capacity of the ultra-lightweight detection head leads to classification collapse. While our 0.14M-parameter R-CNN-Head is sufficient for simpler maritime backgrounds ($R_{0.5} = 0.71$), this weak representation learning capacity proves inadequate for complex urban feature mapping. As evidenced in Table~\ref{Tab_Module}, even for the limited valid proposals successfully forwarded by the RPN ($R_{0.5} = 0.058$), the detection head fails to classify them correctly, resulting in a drastic recall drop to 0.0103. This indicates that the minor parameter constraint directly triggers a training collapse when faced with the difficult classification demands of the UAVID-Tiny dataset.

\subsubsection{Ablation}
Furthermore, we conduct an ablation study of motion information fusion on the R-CNN-Head  using the SeaDroneSee-Tiny dataset. As shown in Table \ref{Tab_Ablation}, explicitly incorporating motion features steadily improves the detection performance across both thresholds ($R_{0.1}$ increases from 0.7306 to 0.7377, and $R_{0.5}$ from 0.7012 to 0.7072) compared to the baseline relying solely on spatial appearance. This confirms that motion dynamics provide crucial complementary cues for accurately distinguishing tiny moving targets from background distractors.

\subsubsection{Hyperparameter Analysis}

Hyperparameter analysis on the validation set of the SeaDroneSee-Tiny dataset is tabulated in Table \ref{Tab_HyperParameter}, focusing on the batch size (BS) and learning rate (LR). The evaluation metric is the $F_2$ score. It presents the $F_2$ scores across combinations of BS in \{16, 32, 64, 128, 256\} and LR ranging from 0.0001 to 0.075. Notably, pairing an excessively high learning rate with a small batch size causes instability; the "N/A" entry at BS = 16 and LR = 0.075 indicates that the training process collapsed. The configuration with BS = 64 and LR = 0.05 achieves the highest F2 score of 0.291. Although the setting with BS = 32 and LR = 0.025 yields the same peak score, we ultimately selected a larger BS as our default configuration because the larger batch size enables faster training. 

\begin{table}[!t]
	\centering
	\caption{Ablation study of motion feature fusion of the R-CNN-Head on SeaDroneSee-Tiny. $R_a$ represents the recall at an NWD threshold of $a$.}
	\label{Tab_Ablation}
	\begin{tabular}{ccc}
		\toprule
		Motion Information & R$_{0.1}$ & R$_{0.5}$ \\
		\midrule
		$\checkmark$ & \textbf{0.7377} &  \textbf{0.7072}  \\
		$\times$& 0.7306 & 0.7012   \\
		\bottomrule
	\end{tabular}%
\end{table}

\subsection{Discussion}

MITE-Net successfully resolves the fundamental dilemma of 4K UAV perception: it avoids both the destructive spatial downsampling of standard YOLO models and the prohibitive latency of slice-aided inference architectures (e.g., SAHI). By deploying the learning-free TTM-RPN as an ultra-fast initial filter, the sub-0.14M-parameter recognition head is spared from dense background computation, processing only salient motion RoIs. This cascaded design validates its viability for resource-constrained edge environments by achieving real-time throughput with exceptional energy efficiency.

However, despite its superior performance in dynamic maritime scenarios, MITE-Net experiences degradation in the hyper-cluttered urban environments of the UAVID-Tiny dataset. This exposes a compound structural bottleneck: the non-learning TTM-RPN struggles to reliably isolate tiny targets from dense urban clutter, while the ultra-lightweight head lacks the representation capacity required to resolve extremely tiny target.

To overcome these limitations, our future research will focus on evolving the fixed-parameter bionic front-end into a fully learnable, end-to-end integrated architecture. By strategically scaling the spatial downsampling factor (e.g., to $ds = 4$ or $8$) to filter macroscopic noise, we can reallocate the FLOP budget to deploy a higher-capacity R-CNN head. Furthermore, we plan to integrate a minimal temporal filter, such as a simple tracker or persistence check, subsequent to the classification head. This addition will effectively reduce the False Alarm Rate (FAR) in complex environments with negligible computational overhead, thereby drastically enhancing the model's generalization while strictly preserving its ultra-low SWaP profile.

\begin{table}[!t]
	\centering
	\caption{Hyperparameter Analysis on the SeaDroneSee-Tiny Validation Set. The table reports the $F_2$ score across various combinations of batch size (BS) and learning rate (LR).}
	\label{Tab_HyperParameter}
	\begin{tabular}{c | c >{\columncolor{blue!10}}c c c c c}
		\toprule
		\diagbox{BS}{LR} & 0.075 & 0.05 & 0.025 & 0.01 & 0.005 & 0.0001 \\
		\midrule
		256 & 0.286 & 0.273 & 0.279 & 0.282 & 0.265 & 0.200 \\
		128 & 0.287 & 0.263 & 0.279 & 0.280 & 0.280 & 0.251 \\
		\rowcolor{blue!10} 64  & 0.246 & \textbf{0.291} & 0.280 & 0.2823 & 0.268 & 0.275 \\
		32  & 0.258    & 0.269 & 0.291  & 0.266 & 0.273 & 0.279 \\
		16  & N/A    & 0.213 & 0.2836  & 0.261 & 0.279 & 0.261 \\
		\bottomrule
	\end{tabular}
\end{table}

\section{Conclusion}\label{Sect_Conclusion}

In this paper, we proposed a comprehensive framework encompassing a novel architecture (MITE-Net), specialized datasets (SAR-Tiny), and hardware-level benchmarks, specifically for 4K tiny target perception in embodied SAR UAVs. By coupling a learning-free bionic motion-perception front-end with an ultra-lightweight R-CNN head ($<0.14$M parameters), the proposed MITE-Net circumvents the computational bottleneck of high-latency full-frame processing, enabling real-time, closed-loop autonomy directly on the edge devices. Benchmarking on the newly constructed and open-sourced SAR-Tiny Datasets, MITE-Net demonstrates exceptional hardware-level efficiency: on an NVIDIA Jetson AGX Xavier, yielding 30.33 FPS and a 100\% search success rate at only 3.19 W (an unprecedented 9.51 FPS/Watt). Furthermore, while extreme tests in hyper-cluttered urban scenes (UAVID-Tiny) successfully mapped the representational boundaries of minimalist heads in static environments, this research establishes a highly efficient and practical baseline for onboard edge intelligence. Moving forward, transitioning the bionic front-end into a fully learnable, end-to-end framework will extend this energy-efficient edge-perception paradigm to the most complex urban environments.

\appendices
\section{Robustness of the Size-Selective Suppression Kernel} \label{App_Size}

\begin{figure*}
	\centering
	\includegraphics[width=1\textwidth]{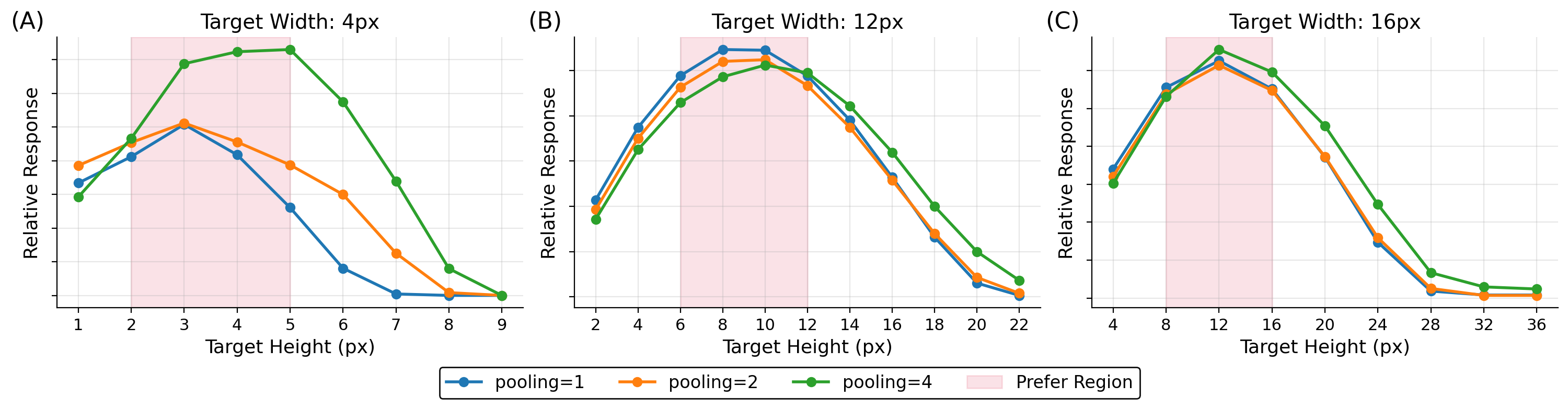}
	\caption{Size selectivity curves of the model for target widths of (A) 4 pixels, (B) 12 pixels, and (C) 16 pixels. By configuring the suppression kernel size to correspond with the input target dimensions, the model consistently achieves its peak relative response when the target height matches the specified target width. The optimal height range yielding this peak response is highlighted by the pink shaded area (Prefer Region). The qualitative trend remains robust across various pooling parameters. Absolute values on the y-axis are omitted to emphasize the relative distribution profile.}
	\label{Fig_Size}
\end{figure*}

To validate the effectiveness of our modified size-selective suppression kernel, we visualize the relative response curves under different target dimensions, as shown in Fig. \ref{Fig_Size}. By adaptively configuring the suppression kernel size according to the input target dimensions, the model is expected to filter out size-mismatched noise. As illustrated, the model consistently achieves its peak relative response when the target height aligns with the specified width (e.g., within the pink shaded "Prefer Region" for widths of 4, 12, and 16 pixels). Most notably, this size-selective behavior remains highly robust and qualitatively consistent across various downsampling scales ($ds \in \{1, 2, 4\}$). This confirms that our adaptive kernel design can effectively and uniformly adjust to specific target sizes, regardless of the variations in feature map resolutions.

\section{Annotation Protocols and Quality Assurance} 
\label{App_CVAT}

To ensure high label fidelity, a five-person team of trained researchers relabeled the datasets using CVAT. The team adhered to strict annotation protocols: (1) \textbf{Tight Bounding:} Boxes must strictly enclose visible target pixels to minimize background bias; (2) \textbf{Temporal Context:} Degraded or occluded targets were only annotated if identifiable using adjacent frames; (3) \textbf{Size Constraints:} Boxes exceeding 1024 pixels were filtered out. 

Furthermore, to mitigate human error inherently associated with tiny objects, we implemented a two-stage Quality Assurance (QA) pipeline. First, a \textbf{Cross-Validation} step was required where a second independent annotator reviewed each batch to flag missed targets or loose boxes. Second, for \textbf{Conflict Resolution}, any discrepancies were escalated to a senior researcher and resolved via consensus, ensuring minimal label noise.

\section{Bounding-Box-Level Comparison} \label{App_Comparison}
\begin{table*}[!t]
	\caption{Performance comparison on SeaDroneSee-Tiny and UAVID-Tiny datasets. Rows with gray-shaded FPS denote theoretical baselines that process raw or sliced 4K imagery to establish accuracy upper bounds, though they fail to meet the real-time requirement ($\ge 30$ FPS) for embodied edge deployment. In contrast, the unshaded rows represent configurations optimized for real-time edge processing.}
	\label{Tab_Overall_Comparison}
	\centering
	\begin{tabular}{c l l c c c c c c c c c r}
		\toprule
		\multirow{2}{*}{\textbf{Dataset}} & \multirow{2}{*}{\textbf{Model}} & \multirow{2}{*}{\textbf{Resolution}} & \multicolumn{3}{c}{\textbf{mAP}} & \multicolumn{3}{c}{\textbf{mAR}} & \multicolumn{3}{c}{\textbf{m$F_2$}} & \multirow{2}{*}{\textbf{FPS}} \\
		\cmidrule(lr){4-6} \cmidrule(lr){7-9} \cmidrule(lr){10-12}
		& & & $et$ & $tiny$ & $all$ & $et$ & $tiny$ & $all$ & $et$ & $tiny$ & $all$ & \\
		\midrule
		\multirow{4}{*}{\parbox{2cm}{\centering SeaDroneSee\\-Tiny}}
		& \multicolumn{1}{>{\columncolor{gray!20}}l}{YOLOv8n-P2 \cite{jocher2023ultralytics}} 
		& \multicolumn{1}{>{\columncolor{gray!20}}l}{SAHI\cite{akyon2022slicing} ($1344\times768\times9$)}
		& \multicolumn{1}{>{\columncolor{gray!20}}c}{0.46}
		& \multicolumn{1}{>{\columncolor{gray!20}}c}{0.59} 
		& \multicolumn{1}{>{\columncolor{gray!20}}c}{0.54} 
		& \multicolumn{1}{>{\columncolor{gray!20}}c}{0.60}
		& \multicolumn{1}{>{\columncolor{gray!20}}c}{0.63} 
		& \multicolumn{1}{>{\columncolor{gray!20}}c}{0.64} 
		& \multicolumn{1}{>{\columncolor{gray!20}}c}{0.60}    
		& \multicolumn{1}{>{\columncolor{gray!20}}c}{0.65}    
		& \multicolumn{1}{>{\columncolor{gray!20}}c}{0.65} 
		& \multicolumn{1}{>{\columncolor{gray!20}}c}{4.79}
		\\
		& YOLOv11n \cite{jocher2024yolo11} & Resize ($1920\times1088$) & 0.24& \textbf{0.46} & \underline{0.18} &0.24& \underline{0.48} &\underline{0.46} &  0.27      &       \textbf{0.48}     &  \textbf{0.46} & 29.15 \\
		& YOLOv8n-P2 \cite{jocher2023ultralytics} & Resize ($1920\times1088$) & \underline{0.26}& \underline{0.42}  &\textbf{0.28} &\underline{0.26}& \underline{0.49}  &\underline{0.47} & \underline{0.28}     &  \underline{0.47}    &  0.37 & \textbf{31.22}\\
		& MITE-Net (Ours) & Raw (4K + 2ds) & \textbf{0.27}& 0.38  &0.08 &\textbf{0.47}& \underline{0.48}  &\textbf{0.48} &   \textbf{0.30}    &   0.40   &  \underline{0.38} & \underline{30.33}\\
		\cmidrule(lr){1-3}
		\multirow{4}{*}{UAVID-Tiny} 
		& \multicolumn{1}{>{\columncolor{gray!20}}l}{YOLOv8n-P2 \cite{jocher2023ultralytics}} 
		& \multicolumn{1}{>{\columncolor{gray!20}}l}{SAHI\cite{akyon2022slicing} ($1344\times768\times9$)}
		& \multicolumn{1}{>{\columncolor{gray!20}}c}{0.04}
		& \multicolumn{1}{>{\columncolor{gray!20}}c}{0.34} 
		& \multicolumn{1}{>{\columncolor{gray!20}}c}{0.16} 
		& \multicolumn{1}{>{\columncolor{gray!20}}c}{0.04}
		& \multicolumn{1}{>{\columncolor{gray!20}}c}{0.47} 
		& \multicolumn{1}{>{\columncolor{gray!20}}c}{0.31} 
		& \multicolumn{1}{>{\columncolor{gray!20}}c}{0.03} 
		& \multicolumn{1}{>{\columncolor{gray!20}}c}{0.31}    
		& \multicolumn{1}{>{\columncolor{gray!20}}c}{0.21}   
		& \multicolumn{1}{>{\columncolor{gray!20}}c}{4.15}
		\\
		& YOLOv11n \cite{jocher2024yolo11} & Resize ($1920\times1088$) & 0.04&  0.10 & 0.04 & 0.03&  0.23 & 0.16 &    0.02  &   0.13   &   0.09  & 30.45 \\
		& YOLOv8n-P2 \cite{jocher2023ultralytics} & Resize ($1920\times1088$) & 0.04&  0.10 & 0.03 & 0.04&  0.23 & 0.17 &   0.02   &  0.14       &   0.10  & 33.17 \\
		& MITE-Net (Ours) & Raw (4K + 2ds) & 0.01 &  0.00  & 0.00  & 0.01 &  0.02  & 0.01  &    0.00   &  0.00    &   0.00 & 31.93\\
		\bottomrule
	\end{tabular}
\end{table*}

As a complement to the embodied SAR evaluation, Table \ref{Tab_Overall_Comparison} details the traditional bounding-box metrics. YOLOv8n-P2 with SAHI establishes the theoretical accuracy upper bound (e.g., 0.54 mAP on SeaDroneSee-Tiny). However, its massive computational overhead (sub-5 FPS) strictly prohibits real-time edge deployment.

Among real-time configurations on SeaDroneSee-Tiny, a clear trade-off emerges. While resized YOLO baselines achieve higher overall mAP (e.g., 0.28), MITE-Net excels in task-specific recall ($mAR_{all} = 0.48$) and dominates the extremely tiny category ($mAP_{et} = 0.27$, $mAR_{et} = 0.47$). This validates MITE-Net's core design philosophy: prioritizing the robust discovery of minuscule targets over exact spatial regression, perfectly correlating with its 100\% SAR search success rate.

Conversely, evaluations on UAVID-Tiny quantitatively expose the model's architectural boundary conditions. In dense urban clutter, while the YOLO baselines also suffer severe degradation (e.g., real-time mAP drops to $\sim$0.04), MITE-Net experiences a near-total representational collapse (near-zero mAP and mAR). Ultimately, this stark collective performance drop underscores the extreme challenge posed by UAVID-Tiny, establishing it as a crucial, rigorous benchmark to drive long-term advancements in complex urban edge-perception.

\section*{Acknowledgment}
The authors employed the Stanford Agentic Reviewer framework\footnote{\url{https://paperreview.ai/}} and Google Gemini\footnote{\url{https://gemini.google.com/}} as supplementary tools to identify potential logical inconsistencies and improve the clarity of the manuscript before submission. All refinements suggested by the tool were critically evaluated and manually implemented by the authors.

\section*{Declaration of Competing Interest}
The authors declare that they have no known competing financial interests or personal relationships that could have appeared to influence the work reported in this paper.

\ifCLASSOPTIONcaptionsoff
  \newpage
\fi

\bibliographystyle{IEEEtran}
\bibliography{MITE_Net_bibfile}

@article{alfarano2024estimating,
  title={Estimating optical flow: A comprehensive review of the state of the art},
  author={Alfarano, Andrea and Maiano, Luca and Papa, Lorenzo and Amerini, Irene},
  journal={Computer Vision and Image Understanding},
  volume={249},
  pages={104160},
  year={2024},
  publisher={Elsevier}
}

@inproceedings{he2017mask,
  title={Mask r-cnn},
  author={He, Kaiming and Gkioxari, Georgia and Doll{\'a}r, Piotr and Girshick, Ross},
  booktitle={Proceedings of the IEEE international conference on computer vision},
  pages={2961--2969},
  year={2017}
}

@software{cvat,
  author       = {{CVAT.ai Corporation}},
  title        = {Computer Vision Annotation Tool (CVAT)},
  month        = jan,
  year         = 2026,
  publisher    = {Zenodo},
  version      = {v2.7.4},
  doi          = {10.5281/zenodo.8416684},
  url          = {https://doi.org/10.5281/zenodo.8416684}
}

@article{howard2017mobilenets,
	title={Mobilenets: Efficient convolutional neural networks for mobile vision applications},
	author={Howard, Andrew G and Zhu, Menglong and Chen, Bo and Kalenichenko, Dmitry and Wang, Weijun and Weyand, Tobias and Andreetto, Marco and Adam, Hartwig},
	journal={arXiv preprint arXiv:1704.04861},
	year={2017}
}

@inproceedings{shang2025centernet,
	title={L-Centernet: A Localization-Based Approach for Aerial Tiny Object Detection},
	author={Shang, Yunqi and Gao, Guangshuai and Dong, Yan},
	booktitle={IGARSS 2025-2025 IEEE International Geoscience and Remote Sensing Symposium},
	pages={8286--8290},
	year={2025},
	organization={IEEE}
}

@article{ma2025owrt,
	title={OWRT-DETR: A novel real-time transformer network for small object detection in open water search and rescue from UAV aerial imagery},
	author={Ma, Shuai and Zhang, Yihong and Peng, Ling and Sun, Chen and Ding, Longbao and Zhu, Yongdong},
	journal={IEEE Transactions on Geoscience and Remote Sensing},
	year={2025},
	publisher={IEEE}
}

@article{liu2019uav,
	title={UAV-assisted wireless powered cooperative mobile edge computing: Joint offloading, CPU control, and trajectory optimization},
	author={Liu, Yuan and Xiong, Ke and Ni, Qiang and Fan, Pingyi and Letaief, Khaled Ben},
	journal={IEEE Internet of Things Journal},
	volume={7},
	number={4},
	pages={2777--2790},
	year={2019},
	publisher={IEEE}
}

@article{deng2020model,
	title={Model compression and hardware acceleration for neural networks: A comprehensive survey},
	author={Deng, Lei and Li, Guoqi and Han, Song and Shi, Luping and Xie, Yuan},
	journal={Proceedings of the IEEE},
	volume={108},
	number={4},
	pages={485--532},
	year={2020},
	publisher={IEEE}
}

@article{sai2023comprehensive,
	title={A comprehensive survey on artificial intelligence for unmanned aerial vehicles},
	author={Sai, Siva and Garg, Akshat and Jhawar, Kartik and Chamola, Vinay and Sikdar, Biplab},
	journal={IEEE Open Journal of Vehicular Technology},
	volume={4},
	pages={713--738},
	year={2023},
	publisher={IEEE}
}

@article{cheng2023towards,
	title={Towards large-scale small object detection: Survey and benchmarks},
	author={Cheng, Gong and Yuan, Xiang and Yao, Xiwen and Yan, Kebing and Zeng, Qinghua and Xie, Xingxing and Han, Junwei},
	journal={IEEE transactions on pattern analysis and machine intelligence},
	volume={45},
	number={11},
	pages={13467--13488},
	year={2023},
	publisher={IEEE}
}

@inproceedings{wang2021tiny,
	title={Tiny object detection in aerial images},
	author={Wang, Jinwang and Yang, Wen and Guo, Haowen and Zhang, Ruixiang and Xia, Gui-Song},
	booktitle={2020 25th international conference on pattern recognition (ICPR)},
	pages={3791--3798},
	year={2021},
	organization={IEEE}
}

@article{erdelj2017help,
	title={Help from the sky: Leveraging UAVs for disaster management},
	author={Erdelj, Milan and Natalizio, Enrico and Chowdhury, Kaushik R and Akyildiz, Ian F},
	journal={IEEE Pervasive Computing},
	volume={16},
	number={1},
	pages={24--32},
	year={2017},
	publisher={IEEE}
}

@article{chen2024unveiling,
	title={Unveiling the power of Haar frequency domain: Advancing small target motion detection in dim light},
	author={Chen, Hao and Sun, Xuelong and Hu, Cheng and Wang, Hongxin and Peng, Jigen},
	journal={Applied Soft Computing},
	volume={167},
	pages={112281},
	year={2024},
	publisher={Elsevier}
}

@article{wang2024BioInspiredSmall,
	title = {Bio-{{Inspired Small Target Motion Detection With Spatio-Temporal Feedback}} in {{Natural Scenes}}},
	author = {Wang, Hongxin and Zhong, Zhiyan and Lei, Fang and Peng, Jigen and Yue, Shigang},
	year = {2024},
	journal = {IEEE Transactions on Image Processing},
	volume = {33},
	pages = {451--465},
	issn = {1057-7149, 1941-0042},
	doi = {10.1109/TIP.2023.3345153},
	urldate = {2024-04-12},
	copyright = {https://ieeexplore.ieee.org/Xplorehelp/downloads/license-information/IEEE.html}
}

@article{Xu2023frac,
	title={A fractional-order visual neural model for small target motion detection},
	author={Xu, Mingshuo and Wang, Hongxin and Chen, Hao and Li, Haiyang and Peng, Jigen},
	journal={Neurocomputing},
	volume={550},
	pages={126459},
	year={2023},
	publisher={Elsevier}
}

@article{ling2022mathematical,
	title={Mathematical study of neural feedback roles in small target motion detection},
	author={Ling, Jun and Wang, Hongxin and Xu, Mingshuo and Chen, Hao and Li, Haiyang and Peng, Jigen},
	journal={Frontiers in Neurorobotics},
	volume={16},
	pages={984430},
	year={2022},
	publisher={Frontiers Media SA}
}

@ARTICLE{Wang2020STMDpuls,  
	author={Wang, Hongxin and Peng, Jigen and Zheng, Xuqiang and Yue, Shigang}, 
	journal={IEEE Transactions on Neural Networks and Learning Systems},   
	title={A Robust Visual System for Small Target Motion Detection Against Cluttered Moving Backgrounds},   
	year={2020},  
	volume={31},  
	number={3},  
	pages={839-853}
}

@article{Wang2022attention,
	title={Attention and Prediction-Guided Motion Detection for Low-Contrast Small Moving Targets},
	author={Wang, Hongxin and Zhao, Jiannan and Wang, Huatian and Hu, Cheng and Peng, Jigen and Yue, Shigang},
	journal={IEEE Transactions on Cybernetics},
	year={2022},
	publisher={IEEE}
}

@article{Wang2021time,
	title={A Time-Delay Feedback Neural Network for Discriminating Small, Fast-Moving Targets in Complex Dynamic Environments},
	author={Wang, Hongxin and Wang, Huatian and Zhao, Jiannan and Hu, Cheng and Peng, Jigen and Yue, Shigang},
	journal={IEEE Transactions on Neural Networks and Learning Systems},
	year={2021},
	publisher={IEEE}
}

@article{Wang2020DSTMD,
	title={A Directionally Selective Small Target Motion Detecting Visual Neural Network in Cluttered Backgrounds}, 
	author={Wang, Hongxin and Peng, Jigen and Yue, Shigang},
	journal={IEEE Transactions on Cybernetics}, 
	year={2020},
	volume={50},
	number={4},
	pages={1541-1555}
}

@article{Wiederman2008ESTMD,
	title={A Model for the Detection of Moving Targets in Visual Clutter Inspired by Insect Physiology},
	author={ Wiederman, S. D  and  Shoemarker, P. A  and  O'Carroll, D. C },
	journal={PLoS ONE},
	volume={3},
	number={7},
	pages={e2784-},
	year={2008}
}

@article{Nordstrom2006small,
	title={Small object detection neurons in female hoverflies},
	author={Nordstr{\"o}m, Karin and O'Carroll, David C},
	journal={Proceedings of the Royal Society B: Biological Sciences},
	volume={273},
	number={1591},
	pages={1211--1216},
	year={2006},
	publisher={The Royal Society London}
}

@article{Nordstrom2006insect,
	title={Insect detection of small targets moving in visual clutter},
	author={Nordstr{\"o}m, Karin and Barnett, Paul D and O'Carroll, David C},
	journal={PLoS biology},
	volume={4},
	number={3},
	pages={e54},
	year={2006},
	publisher={Public Library of Science San Francisco, USA}
}

@article{Barnett2007retinotopic,
	title={Retinotopic organization of small-field-target-detecting neurons in the insect visual system},
	author={Barnett, Paul D and Nordstr{\"o}m, Karin and O'carroll, David C},
	journal={Current Biology},
	volume={17},
	number={7},
	pages={569--578},
	year={2007},
	publisher={Elsevier}
}

@article{Nordstrom2012neural,
	title={Neural specializations for small target detection in insects},
	author={Nordstr{\"o}m, Karin},
	journal={Current opinion in neurobiology},
	volume={22},
	number={2},
	pages={272--278},
	year={2012},
	publisher={Elsevier}
}

@software{jocher2024yolo11,
	author = {Glenn Jocher and Jing Qiu},
	title = {Ultralytics YOLO11},
	version = {11.0.0},
	year = {2024},
	url = {https://github.com/ultralytics/ultralytics},
	license = {AGPL-3.0}
}

@software{jocher2023ultralytics,
	author = {Glenn Jocher and Ayush Chaurasia and Jing Qiu},
	title = {Ultralytics YOLO},
	version = {8.0.0},
	year = {2023},
	url = {https://github.com/ultralytics/ultralytics},
	license = {AGPL-3.0}
}

@article{lecun2015deep,
	title={Deep learning},
	author={LeCun, Yann and Bengio, Yoshua and Hinton, Geoffrey},
	journal={nature},
	volume={521},
	number={7553},
	pages={436--444},
	year={2015},
	publisher={Nature Publishing Group UK London}
}

@inproceedings{liu2016ssd,
	title={Ssd: Single shot multibox detector},
	author={Liu, Wei and Anguelov, Dragomir and Erhan, Dumitru and Szegedy, Christian and Reed, Scott and Fu, Cheng-Yang and Berg, Alexander C},
	booktitle={European conference on computer vision},
	pages={21--37},
	year={2016},
	organization={Springer}
}

@article{wang2021normalized,
	title={A normalized Gaussian Wasserstein distance for tiny object detection},
	author={Wang, Jinwang and Xu, Chang and Yang, Wen and Yu, Lei},
	journal={arXiv preprint arXiv:2110.13389},
	year={2021}
}

@inproceedings{jiang2018acquisition,
	title={Acquisition of localization confidence for accurate object detection},
	author={Jiang, Borui and Luo, Ruixuan and Mao, Jiayuan and Xiao, Tete and Jiang, Yuning},
	booktitle={Proceedings of the European conference on computer vision (ECCV)},
	pages={784--799},
	year={2018}
}

@article{li2020generalized,
	title={Generalized focal loss: Learning qualified and distributed bounding boxes for dense object detection},
	author={Li, Xiang and Wang, Wenhai and Wu, Lijun and Chen, Shuo and Hu, Xiaolin and Li, Jun and Tang, Jinhui and Yang, Jian},
	journal={Advances in neural information processing systems},
	volume={33},
	pages={21002--21012},
	year={2020}
}

@article{ge2021yolox,
	title={Yolox: Exceeding yolo series in 2021},
	author={Ge, Zheng and Liu, Songtao and Wang, Feng and Li, Zeming and Sun, Jian},
	journal={arXiv preprint arXiv:2107.08430},
	year={2021}
}

@article{liu2018intriguing,
	title={An intriguing failing of convolutional neural networks and the coordconv solution},
	author={Liu, Rosanne and Lehman, Joel and Molino, Piero and Petroski Such, Felipe and Frank, Eric and Sergeev, Alex and Yosinski, Jason},
	journal={Advances in neural information processing systems},
	volume={31},
	year={2018}
}

@inproceedings{yu2017dilated,
	title={Dilated residual networks},
	author={Yu, Fisher and Koltun, Vladlen and Funkhouser, Thomas},
	booktitle={Proceedings of the IEEE conference on computer vision and pattern recognition},
	pages={472--480},
	year={2017}
}

@article{ren2015faster,
	title={Faster {r-cnn}: Towards real-time object detection with region proposal networks},
	author={Ren, Shaoqing and He, Kaiming and Girshick, Ross and Sun, Jian},
	journal={Advances in neural information processing systems},
	volume={28},
	year={2015}
}

@article{ren2016faster,
	title={Faster R-CNN: Towards real-time object detection with region proposal networks},
	author={Ren, Shaoqing and He, Kaiming and Girshick, Ross and Sun, Jian},
	journal={IEEE transactions on pattern analysis and machine intelligence},
	volume={39},
	number={6},
	pages={1137--1149},
	year={2016},
	publisher={IEEE}
}

@inproceedings{akyon2022slicing,
	title={Slicing aided hyper inference and fine-tuning for small object detection},
	author={Akyon, Fatih Cagatay and Altinuc, Sinan Onur and Temizel, Alptekin},
	booktitle={2022 IEEE international conference on image processing (ICIP)},
	pages={966--970},
	year={2022},
	organization={IEEE}
}

@inproceedings{redmon2016you,
	title={You only look once: Unified, real-time object detection},
	author={Redmon, Joseph and Divvala, Santosh and Girshick, Ross and Farhadi, Ali},
	booktitle={Proceedings of the IEEE conference on computer vision and pattern recognition},
	pages={779--788},
	year={2016}
}

@article{ying2025VisibleThermalTiny,
  title = {Visible-{{Thermal Tiny Object Detection}}: {{A Benchmark Dataset}} and {{Baselines}}},
  shorttitle = {Visible-{{Thermal Tiny Object Detection}}},
  author = {Ying, Xinyi and Xiao, Chao and An, Wei and Li, Ruojing and He, Xu and Li, Boyang and Cao, Xu and Li, Zhaoxu and Wang, Yingqian and Hu, Mingyuan and Xu, Qingyu and Lin, Zaiping and Li, Miao and Zhou, Shilin and Sheng, Weidong and Liu, Li},
  year = {2025},
  journal = {IEEE Transactions on Pattern Analysis and Machine Intelligence},
  pages = {1--8},
  issn = {1939-3539},
  doi = {10.1109/TPAMI.2025.3544621},
  urldate = {2025-05-17}
}

@article{kiefer20221st, 
title={1st Workshop on Maritime Computer Vision (MaCVi) 2023: Challenge Results}, 
author={Kiefer, Benjamin and Kristan, Matej and Per{\v{s}}, Janez and {\v{Z}}ust, Lojze and Poiesi, Fabio and Andrade, Fabio Augusto de Alcantara and Bernardino, Alexandre and Dawkins, Matthew and Raitoharju, Jenni and Quan, Yitong and others}, 
journal={arXiv preprint arXiv:2211.13508}, 
year={2022} 
}

@inproceedings{varga2022seadronessee,
title={Seadronessee: A maritime benchmark for detecting humans in open water},
author={Varga, Leon Amadeus and Kiefer, Benjamin and Messmer, Martin and Zell, Andreas},
booktitle={Proceedings of the IEEE/CVF Winter Conference on Applications of Computer Vision},
pages={2260--2270},
year={2022} 
}

@article{lyu2020uavid,
  title={UAVid: A semantic segmentation dataset for UAV imagery},
  author={Lyu, Ye and Vosselman, George and Xia, Gui-Song and Yilmaz, Alper and Yang, Michael Ying},
  journal={ISPRS journal of photogrammetry and remote sensing},
  volume={165},
  pages={108--119},
  year={2020},
  publisher={Elsevier}
}

@article{xu2025vstmd,
  title={vSTMD: Visual Motion Detection for Extremely Tiny Target at Various Velocities},
  author={Xu, Mingshuo and Luan, Hao and Hao, Zhou Daniel and Peng, Jigen and Yue, Shigang},
  journal={arXiv preprint arXiv:2501.13054},
  year={2025}
}

\end{document}